\documentclass{article}

\usepackage{arxiv}

\usepackage[utf8]{inputenc} % allow utf-8 input
\usepackage[T1]{fontenc}    % use 8-bit T1 fonts
\usepackage{hyperref}       % hyperlinks
\usepackage{url}            % simple URL typesetting
\usepackage{booktabs}       % professional-quality tables
\usepackage{amsfonts}       % blackboard math symbols
\usepackage{nicefrac}       % compact symbols for 1/2, etc.
\usepackage{microtype}      % microtypography
\usepackage{lipsum}		% Can be removed after putting your text content
\usepackage{graphicx}
\usepackage{doi}
\usepackage{amssymb}
\usepackage{graphicx}
\usepackage{epsfig}
\usepackage{amsmath}
\usepackage{floatrow}
\usepackage{enumitem}
\usepackage{bm}
\usepackage{mathtools}
\newcommand{\specialcell}[2][c]{%
\begin{tabular}[#1]{@{}c@{}}#2\end{tabular}}

\title{Unsupervised Legendre-Galerkin Neural Network for Singularly Perturbed Partial Differential Equations}

\author{ {\hspace{1mm}Junho Choi} \\
	Department of Mathematics\\
         Sungkyunkwan University\\
	Suwon, South Korea \\
	\texttt{junho.choi@skku.edu} \\
	\And
	{\hspace{1mm}Namjung Kim} \\
	Department of Mechanical Engineering\\
          Gachon University\\
	Seongnam, South Korea \\
	\texttt{namjungk@gachon.ac.kr} \\
	\And
	{\hspace{1mm}Youngjoon Hong} \\
	Department of Mathematics\\
         Sungkyunkwan University\\
	Suwon, South Korea \\
	\texttt{hongyj@skku.edu}
}

\renewcommand{\shorttitle}{Unsupervised Legendre-Galerkin Neural Network }

\hypersetup{
pdftitle={Unsupervised Legendre-Galerkin Neural Network for Singularly Perturbed Partial Differential Equations},
pdfauthor={Choi. JH, Kim. NJ, Hong. YJ},
pdfkeywords={unsupervised learning, 
deep neural network, 
Legendre-Galerkin approximation,
spectral bias, 
boundary layer, 
singular perturbation.}
}

\begin{document}
\maketitle

\begin{abstract}
	Machine learning methods have been lately used to solve partial differential equations (PDEs) and dynamical systems.
These approaches have been developed into a novel research field known as scientific machine learning in which techniques such as deep neural networks and statistical learning are applied to classical problems of applied mathematics.
In this paper, we develop a novel numerical algorithm that incorporates machine learning and artificial intelligence to solve PDEs.
Based on the Legendre-Galerkin framework, we propose the {\it unsupervised machine learning} algorithm to learn {\it multiple instances} of the solutions for different types of PDEs. Our approach overcomes the limitations of data-driven and physics-based methods.
The proposed neural network is applied to general 1D and 2D PDEs with various boundary conditions as well as convection-dominated {\it singularly perturbed PDEs} that exhibit strong boundary layer behavior.
\end{abstract}

% keywords can be removed
\keywords{unsupervised learning, 
deep neural network, 
Legendre-Galerkin approximation,
spectral bias, 
boundary layer, 
singular perturbation.}

\section{Introduction}
\label{intro}
Modern machine learning methods using deep neural networks (DNNs) have demonstrated outstanding performance in image recognition \cite{krizhevsky2012imagenet}, natural language processes \cite{lecun2015deep}, cognitive science \cite{lake2015human}, and time series classification \cite{cui2016multi}.
Lately, machine learning methods have been widely used to solve partial differential equations (PDEs) and dynamical systems.
Since neural networks are capable of approximating nonlinear functions, computational parameterization through machine learning and optimization methods has shown noticeable performance to numerically solve various PDEs.
As a result, these approaches have opened a new fields of research known as scientific machine learning.
In this paper, we provide a novel algorithm in the field of numerical method for PDEs using machine learning and artificial intelligence.
More precisely, we propose an unsupervised machine learning algorithm based on the Legendre-Galerkin neural network framework to find an accurate approximation to the solution of different types of PDEs.
It is noteworthy that our method can learn multiple instances without training database.
The proposed neural network is applied not only to general 1D and 2D differential equations but also to convection-domindated singularly perturbed PDEs that exhibit boundary layer behavior.

Many neural network approaches without solution datasets (training datasets) have been successfully developed, such as physics informed neural networks (PINN) \cite{blechschmidt2021three,kharazmi2019variational,han2021hierarchical}, deep Ritz method (DRM) \cite{yu2018deep}, and deep Galerkin method (DGM) \cite{sirignano2018dgm}.
Instead of using the solution dataset, these approaches introduce a residual function to define a loss function derived from differential equations.
In particular, PINN uses collocation points as training data points in the space-time domain \cite{PINN001, PINN002,PINN003,PINN004, PINN005, PINN007, PINN008, xu2021weak, kharazmi2019variational}.
Hence, the PINN algorithm is available for time dependent, multi-dimensional PDEs, and computational domains of various shapes; see e.g., \cite{zhang2019quantifying, wang2021eigenvector, wang2022fractional, yu2022gradient, wu2023comprehensive}. %{\color{blue}In addition, the PINN is extended to various kinds of methods, for example, a method for stochastic PDEs in \cite{zhang2019quantifying}, for multi-scale PDEs in \cite{wang2021eigenvector}, for fractional PDEs in \cite{wang2022fractional}, a gradient-enhanced method in \cite{yu2022gradient}, and a method on sampling points by residual-based adaptation in \cite{wu2023comprehensive}.}
However, the neural networks optimize only a single instance from a fixed input, such as initial conditions, boundary conditions, external forcing terms, and coefficients.
Therefore, when the input is changed, the training process has to be repeated. 
Obtaining outputs in real time for complex systems that require various sets of input data is difficult with these approaches that learn a single PDE solution.
On the other hand, data-supervised or data-driven methods (DDM) have been explored in \cite{bar2019learning, zhuang2021learned, li2020fourier, chudomelka2020deep,rudy2017data, xiu01, xiu02, xiu03}. 
Since the loss functions of DDMs directly compare a true solution with an approximation generated by DNNs, neural networks fit the numerical solution in a process similar to supervised learning tasks in machine learning.
Since neural networks for DDMs are designed to take various input data such as initial conditions (ICs), boundary conditions (BCs), and external forcing terms, they are able to produce multiple instances of the PDEs.
However, to train DNNs, solutions of PDEs are required as a training dataset, which must be analytically or numerically provided in advance. 
The deep operator network (DeepONet, \cite{lu2021learning}) has been also robustly studied lately in multiple fields, such as, multiphysics and multiscale problems in \cite{cai2021deepm,mao2021deepm}, a method using Fourier Neural Operator in \cite{lu2022comprehensive}, a method for multiple-input operators in \cite{jin2022mionet}, and a method for input data with various degrees of accuracy in \cite{lu2022multifidelity}. 
The authors in \cite{PIDON, PINO} introduced an operator learning approach that can learn multiple PDE solutions. However, as in other ML based approaches, they are not able to accurately predict numerical solutions to the stiff PDEs which exhibit a sharp transition near the boundary.
Recently, Galerkin Neural Network (GNN) was introduced to learn a test function for a single instance in \cite{ainsworth2021galerkin,ainsworth2022galerkin} and solve a boundary layer issue arose from reaction-diffusion types of singular perturbation problems.

In this paper, we develop an unsupervised Legendre Galerkin neural network (ULGNet) to solve various (partial) differential equations without the training dataset consisting of solutions of the PDE.
When the BCs and external forcing functions are given as inputs, the proposed ULGNet predicts a numerical solutions to the PDEs.
Hence, the ULGNet can learn multiple instances of the solutions of PDEs.
The loss function is set as a residual quantity motivated by the weak formulation of the Legendre-Galerkin approximation, as in the spectral element methods \cite{shen2011spectral}.
Generating the training dataset is often inefficient because the process relies on numerical methods such as finite difference, finite element, and finite volume approximation. 
%Indeed, a large quantity of numerical solutions have to be obtained by using massive numerical computations in order to construct a reliable solution dataset.
Although unsupervised learning seems efficient, most unsupervised algorithms for scientific machine learning predict only a single instance of the PDE. Hence, when the input information of initial conditions, boundary condition, and external force is changed, the neural network should be retrained. 
The proposed ULGNet is designed to overcome the drawbacks of recently developed algorithms in the scientific machine learning.
The ULGNet does not need to generate the solution dataset, as it is {\it unsupervised learning}, as well as the algorithm predicts {\it multiple instances} of PDE solutions. 
In order to construct an approximation to the solution of PDEs, we borrow a framework from the spectral method (SM) as in \cite{chudomelka2020deep}. 
In particular, once the ULGNet infers coefficients, say $\alpha_i$, of polynomial basis functions $\phi_i$, we construct the linear combination $\sum_{i=0}^{N-1}\alpha_i\phi_i$ to approximate a solution to PDEs \cite{gottlieb1977numerical,shen2011spectral}. 
Since each basis function satisfies the exact boundary condition, the predicted solution obtains the exact boundary condition.
In addition, since the SM is known to be a high resolution numerical scheme, we expect our predicted numerical solution to yield relatively small errors compared to other machine learning based approaches.

The proposed ULGNet consists of a convolutional neural network (CNN) and a nonlinear activation function between each layer. 
A convolutional layer not only compresses the input data and maintains correlation between the input. 
Hence, the main architecture of the ULGNets is adopted from network structures used in the computer vision tasks \cite{cui2016multi,krizhevsky2012imagenet}. 
We consider the Legendre-Galerkin (LG) framework of the boundary value problem at the heart of our full solver.
Compared to the residual of strong form of PDEs in PINNs, the weak form can reduce higher order differentiation to lower order by integration by parts to lessen the errors caused by higher order differentiation. 
Hence, the performance of the network is further improved by
avoiding the high order numerical derivatives.
In addition, we set the Legendre polynomial as a test function of the LG approximation in the simulations.
Hence, the numerical integration is highly accurate due to Gauss–Legendre quadrature.

%%%%%%%%%
%%%%%%%%%
%%%%%%%%%

One of the main contributions of our study is the development of a learning architecture that accurately solve singular perturbation and boundary layer problems.
Numerical methods for singularly perturbed convection-diffusion problems pose the substantial difficulties since a small diffusive parameter produces a sharp transition inside thin layers. 
In most classical simulations of singularly perturbed equations, the computed solutions produce a large numerical errors when the diffusive term is small.
Hence, semi-analytic methods have been proposed and successfully applied as “enriched spaces”. 
The main component of the enriched space method is to add a global basis function (a so-called corrector function in the analysis) and supplement the discrete space with the corrector function.
The corrector functions constructed by singular perturbation analysis can absorb the boundary/interior layer singularities near the sharp transition.
In machine learning, it is well-known that spectral bases prioritize learning of low-frequency components of the target function \cite{rahaman2019spectral}.
More precisely, since neural networks rely on the smooth prior, the spectral bias leads to a failure to accurately capture high oscillation or sharp transition in solution functions.
Hence, without care, neural networks cannot capture the sharp transition caused by the boundary layer \cite{wang2022and, fbpinns}.
In our algorithm, since the coefficients of the spectral approximation are predicted by the neural network, the main structure of the numerical scheme is maintained. 
From this perspective, existing numerical methods, such as enriched space
methods, are adoptable to the ULGNet architecture by adding a proper basis function according to \cite{hong2018enriched,jung2005numerical, hong2014numerical, H2020}; for more details, see e.g. Section 4.
Hence, the ULGNet successfully resolve the boundary layer behavior which is not well-studied in recent machine learning approaches.
In Table \ref{t:01}, we provide a comparison of recent machine learning approaches for solving PDEs.

Our contributions are listed as follows:
\begin{itemize}
    \item We propose a novel ULGNet architecture based on the Legendre-Galerkin framework, which learns multiple instances of solutions without the training dataset.
    \item The predicted solution inferred by the proposed algorithm satisfies an exact boundary condition, which alleviates numerical errors from boundary values.
    \item The ULGNet expects low generalization errors since the structure of our basis framework enhances expressive power of our algorithm.
    \item The ULGNet provides a learning architecture that accurately solve convection-dominated singular perturbed problems that exhibit strong boundary layer.
\end{itemize}

 \begin{table}[t!]  
  \begin{tabular}{ |c|c|c|c| } 
 \hline
 & {Training dataset} & PDE instances & Enriched scheme
  \\ \hline
  PINN, DRM, DGM & not required & single instance & unavailable
  \\ \hline
  FNO, DON, DDM & required & multiple instances & unavailable
  \\ \hline
  PIDON, PINO & not required & multiple instances & unavailable
  \\ \hline
  ULGNet (ours) & not required & multiple instances & available
  \\ \hline
\end{tabular}
%  \begin{tabular}{ |c|c|c|c|c| } 
%  \hline
%   & PINN, DRM, DGM & FNO, DON, DDM & PIDON &ULGNet (ours) \\ \hline
%   Training dataset & not required &  required & not required  & not required\\  \hline
  
%  PDE instances   & single  & multiple  & multiple  & multiple  \\ \hline
%   Enriched scheme   & unavailable & unavailable & unavailable & available \\ \hline
% \end{tabular}
\caption{Comparison of machine learning approaches for solving PDEs.
Here, PINN, DRM, DGM, FNO, DON, DDM, PINO, and PIDON stand for Physics Informed Neural Network \cite{PINN001}, Deep Ritz Method \cite{yu2018deep}, Deep Galerkin Method \cite{sirignano2018dgm}, Fourier Neural Operator \cite{li2020fourier}, Deep Operator Network \cite{lu2021learning}, Data Driven Discretization \cite{bar2019learning}, physics-informed neural operator \cite{PINO}, and physics-informed DeepONet \cite{PIDON}, respectively.}
\label{t:01}
\end{table}

%The outline of this article is as follows. In Section \ref{method}, we describe the architecture of the ULGNet and introduce our loss function based on the  weak formulation. In Sections \ref{sec:3}, training on various types of PDEs are implemented including two-dimensional problems.  In Section \ref{sec:4}, we demonstrate how the proposed ULGNets with a loss function handle singularly perturbed convection-dominated equations. We conclude the paper with a summary in Section \ref{discussion}.

\section{Method}
\label{method}
Since the ULGNet is motivated by the spectral element method, we briefly describe the LG spectral element method to guide our algorithm.
Let us consider second-order elliptic differential equations with boundary conditions
\begin{align}\label{elliptic}
\begin{split}
{\color{black}-\epsilon\Delta u}+\mathcal{F}(u,\nabla u)=f(x),\quad & x\in\Omega\subset \mathbb{R}^n,\\
\mathcal{B}(u,\nabla u)=0,\quad \text{at}\quad & x\in\partial \Omega,
\end{split}
\end{align}
where $\mathcal{F}$ is linear or nonlinear operator, $\mathcal{B}$ is a boundary operator, and $\epsilon  > 0$ is a constant.
An integral form or a weak form of \eqref{elliptic} can be defined as 
\begin{align}\label{weak_form}
\int_\Omega {\color{black}\epsilon\nabla u\cdot\nabla\phi}+\mathcal{F}(u,\nabla u)\phi dx=\int_\Omega f\phi dx
\end{align}
for any $\phi$ in an appropriate Hilbert space $H$. 
Thanks to the elliptic theory   \cite{evans2022partial,gilbarg1977elliptic,krylov2008lectures}, it is well-known that there exists a weak solution to satisfy \eqref{weak_form} in $H$. 
The numerical solution of \eqref{weak_form} in $H$ can be represented as
\begin{align}\label{representation}
u(x)=\sum_{k=0}^{\infty}\alpha_k\phi_k(x).
\end{align}
Throughout this paper, we set the basis function to be a compact combination of Legendre polynomials such as
\begin{align}\label{basis}
\phi_k(x)=L_k(x)+a_kL_{k+1}(x)+b_kL_{k+2}(x),\quad x\in [-1,1]
\end{align}
where $L_k$ are Legendre polynomials of degree $k$. 
One can represent various boundary conditions by choosing a proper $a_k$ and $b_k$; see e.g., \cite{shen2011spectral}. 
This paper provides examples for the homogeneous Dirichlet boundary (see \eqref{basis_dirichlet} for more details) condition as well as the homogeneous Neumann boundary condition (see \eqref{basis_neumann}) as a paradigm example. Apply the LG method to \eqref{representation}, we find an approximation to PDEs in a finite subspace of $H$ spanned by $\{\phi_k\}_{k=0}^{N-2}$ where $N$ is a finite integer, that is, 
\begin{align}\label{appox}
u(x)\approx\sum_{k=0}^{N-2}\alpha_k\phi_k(x).
\end{align}

\begin{figure}[t]
     \includegraphics[width=14cm]{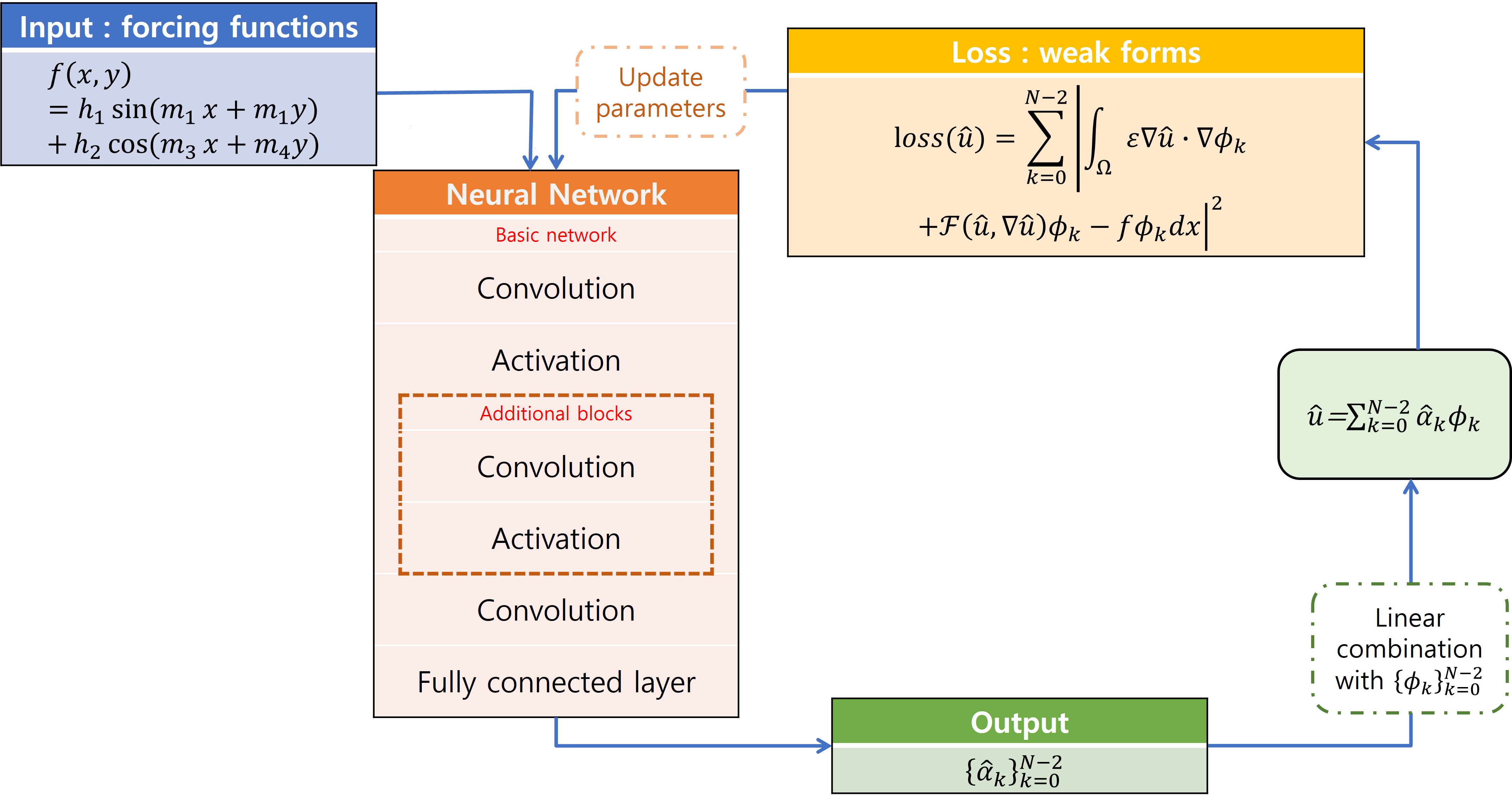}
\caption{Schematic diagram of the unsupervised
Legendre-Galerkin Deep Neural Network(ULGNet).}
\label{architecture}
\end{figure}
Now, the LG method aims to determine the coefficients $\alpha_k$ for $0 \leq k \leq N-2$. 
While the spectral method is quiet similar to the finite elements method, the basis functions in the spectral method are global and continuously differentiable on the whole domain. 
Specifically, if a function $u$ is in a Hilbert space, $H^{m}$, then the corresponding approximation $\sum_{k=0}^{N-1}\alpha_k\phi_k$ converges to $u$ within $O(N^{-m})$; for more details see e.g., \cite{shen2011spectral}.
Hence, provided that $u$ is smooth enough, the spectral method have excellent error properties, that is, it achieves the so-called exponential convergence.
The compact form can represent various boundary conditions such as the Dirichlet (see Section \ref{CDE_sect}) and Neumann conditions (see Section \ref{HE_sect}) by choosing $a_k$ and $b_k$. 
In addition, the orthogonality of the Legendre polynomials enables the mass and stiff matrices to be sparse; see e.g.,  {\cite{shen2011spectral}. }

Regarding the neural network architecture, we construct ULGNet based on the Legendre-Galerkin Deep Neural Network (LGNet) algorithm introduced in our previous study \cite{chudomelka2020deep}. 
The input of the neural network is the external forcing function of the PDEs.
Then, the input feature passes through the neural network and produces an output,  $\{\widehat{\alpha}_k\}$. 
After that, $\widehat{u}$ is reconstructed by
\begin{align}\label{inference}
\widehat{u}(x)\approx\sum_{k=0}^{N-2}\widehat{\alpha}_k\phi_k(x).
\end{align}
In order to train the ULGNet without a true solution, we employ a weak form as a loss function, as given below. 
\begin{align}\label{loss}
loss(\widehat{u}):=\sum_{k=0}^{N-2}\left|\int_\Omega{\color{black}\epsilon\nabla \widehat{u}\cdot\nabla\phi_k}+\mathcal{F}(\widehat{u},\nabla \widehat{u})\phi_k - f\phi_k dx\right|^2.
\end{align}
The parameters in the ULGNet should be updated to minimize the loss function defined in \eqref{loss}.
For the optimization algorithm, we adopt the limited-memory Broyden–Fletcher–Goldfarb–Shanno (L-BFGS) method, as in \cite{liu1989limited}. 
If the number of Legendre-Gauss-Lobatto (LGL) collocation points is $N+1$, the number of corresponding basis functions is $N-1$. Hence, the loss function in \eqref{loss} uses $N-1$ test functions in total. 
Figure \ref{architecture} displays a schematic diagram of our method, the ULGNet. Once forcing functions $f$ are given as input data, they are inserted into the ULGNet. Subsequently, the input data pass the neural network which consists of multiple layers of convolution and activation. 
Then, the feature is flattened by a fully connected network. 
For the output, a set of coefficients $\{\widehat{\alpha}_k\}_{k=0}^{N-2}$ of the Legendre polynomials $\{\phi_k\}_{k=0}^{N-2}$ is obtained. 
The approximations $\widehat{u}$ corresponding to the given forcing function $f$ are constructed by $\sum_{k=0}^{N-2}\widehat{\alpha}_k\phi_k$. After that, the loss of $\widehat{u}$ are measured by $N$ weak formulations of the target PDEs in order to compute the difference between $\widehat{u}$ and the true solution in a certain norm.

To close this section, we define some metrics below to measure the errors. 
The relative $L^2$ norm is defined as
\begin{align}\label{rel_l2}
Rel.L^2(u,\widehat{u}):= \sqrt{\frac{\int_{-1}^1 |u-\widehat{u}|^2dx}{\int_{-1}^1 |{u}|^2dx}}.
\end{align}
The mean absolute error (MAE) is written as
\begin{align}\label{MAE}
MAE(u,\widehat{u}):= \frac{1}{N+1}\sum_{i=0}^{N}|u_i-\widehat{u}_i|,
\end{align}
where $N$ is the number of nodal points.
The $L^\infty$ norm is defined by
\begin{align}\label{Linfty}
L^\infty(u,\widehat{u}):= \max_{x\in\Omega}|u(x)-\widehat{u}(x)|.
\end{align}

\section{Numerical experiments: non-stiff differential equations}
\label{sec:3}
We train an ULGNet model to solve non-stiff (partial) differential equations. 
Since the solution profile is smooth, the LG framework can successfully approximate the solution. 
In Table \ref{overview}, convergence and useful hyperparameters are presented to demonstrate the performance of our proposed algorithm.
\subsection{Convection Diffusion Equation}\label{CDE_sect}
As a paradigm example, we demonstrate that the ULGNet is able to solve the one-dimensional convection diffusion equation (CDE) with the homogeneous Dirichlet boundary condition:
\begin{align}\label{CDE}
\begin{split}
&-\epsilon u_{xx}-u_x=f,\quad x\in (-1,1)=:\Omega,\\
&u(-1)=0=u(1),
\end{split}
\end{align}
where $\epsilon=0.1$. 
\begin{table}[t!]
\begin{center}
\resizebox{\columnwidth}{!}{
\begin{tabular}{||c c c c c c c ||} 
 \hline
 Equation & BC & Blocks& \specialcell{Number \\of inputs} & Epochs & \specialcell{PDE\\ coefficient}&  \specialcell{Mean\\ Rel. $L^2$\\ error}   \\ 
 \hline\hline
 CDE \eqref{CDE} & Dirichlet & \specialcell{basic network \\with ReLU} & 10000& $5\cdot10^4$ &$\epsilon = 10^{-1}
 $& $8.151\cdot10^{-4}$ \\ 
 \hline
Helmholtz \eqref{HE} & Neumann & \specialcell{one block \\with ReLU}  & 10000& $10^5$ &$k_u = 3.5$& $3.906\cdot10^{-3}$ \\
 \hline
Burgers \eqref{BE} & Dirichlet &  \specialcell{one block\\ with Swish} & 10000& $10^5$ &$\epsilon = 10^{-1}
 $& $2.787\cdot10^{-3}$ \\
 \hline
2D CDE \eqref{2D_CDE} & Dirichlet & \specialcell{basic network  \\with ReLU} & 5000& $10^5$ &$\epsilon = 10^{-1}
 $& $4.446\cdot10^{-3}$ \\
 \hline
 \specialcell{1D CDE with\\
 a boundary layer \eqref{CDEB}} & Dirichlet & \specialcell{one block \\with ReLU} & 10000& $5\cdot10^4$ &$\epsilon = 10^{-6}
 $& $7.354\cdot10^{-4}$ \\ \hline 
\specialcell{2D CDE with\\
a boundary layer \eqref{2D_CDEB}} & Dirichlet & \specialcell{one block \\with ReLU} & 8000& $2\cdot10^4$ &$\epsilon = 10^{-6}
 $& $7.369\cdot10^{-4}$ \\ \hline
\end{tabular}
}
\end{center}
\caption{Numerical convergence of the ULGNet.}\label{overview}
\end{table}
\label{experiments}
%{\color{blue}It is well known that if $f\in C^{m}(\Omega)$, there is a weak solution of \eqref{weak_form} in $H_0^{m+2}(\Omega)$ for some positive integer $m$. Moreover, basis functions of $H_0^{m+2}(\Omega)$ can be constructed with the Legendre polynomials as follows.} 
To assign the homogeneous boundary condition, we set $a_k = 0$ and $b_k = -1$ for all $k$ in \eqref{basis}.
Hence, the compact form of the Legendre polynomial basis can be written as 
\begin{align}\label{basis_dirichlet}
\phi_k(x)=L_k(x)-L_{k+2}(x),\quad x\in [-1,1],
\end{align}
where $L_k$ is the $k$-th Legendre polynomial.  
In our experiments, we choose $N+1=32$ for the LGL nodal points. Then, $\{\phi_k\}_{k=0}^{29}$ becomes the corresponding set of basis functions for the nodal points. 
%{\color{blue}Therein, a subspace of $H_0^{m+2}(\Omega)$ spanned by $\{\phi_k\}_{k=0}^{29}$ is where the ULGNet yields approximations to the weak solutions.} 
Following \cite{chudomelka2020deep}, we use the ReLU activation function for the linear problem with 32 convolutional filters and a kernel size of 5.    

As for inputs of the ULGNet, we generate $P=10,000$ forcing functions which are given by  
\begin{align}\label{EF}
f_i(x)=h_{1i}\sin(m_{1i}x)+h_{2i}\cos( m_{2i}x),
\end{align}
where $h_{ji}$ and $m_{ji}$ for $j=1,2$ and $i=1,2,\cdots,P$ are drawn from a uniform distribution on $[3,5]$ and $[0, 2 \pi]$, respectively.
After feeding the inputs, the ULGNet yields 10,000 sets of the coefficients $\{\alpha_{ik}\}_{k=0}^{29}$ corresponding to the i-th forcing function, $f_i$.
As a result, the i-th predicted numerical solutions $\widehat{u}_i$ is constructed by \eqref{inference}. We define a loss function derived from weak formulations of \eqref{CDE} as: 
\begin{align} \label{CDE_weak}
\epsilon \int_{-1}^1 (u_i)_{x}(\phi_j)_x dx-\int_{-1}^1 (u_i)_{x}\phi_jdx=\int_{-1}^1 f_i\phi_jdx,
\end{align}
for $0 \leq j \leq 29$. 
From the weak formulations, the loss function for the ULGNet is defined as
\begin{align}\label{weak_loss}
loss=\sum_{i=1}^{P}\sum_{j=0}^{N-2}\left|\int_{-1}^1\epsilon (\widehat{u}_i)_x(\phi_j)_x-(\widehat{u}_i)_x\phi_j- f_i\phi_jdx\right|^2,
\end{align}
where $\widehat{u}_i$ is the predicted solution of $u_i$, and $P$ is the number of inputs. The trained neural network is then tested using an out-of-sample set (unseen inputs) of $1,000$ randomly generated functions as in \eqref{EF}. These 1,000 inputs for testing are not present in the inputs for training.

\begin{figure}[t]
\begin{tabular}{cc}
     \includegraphics[width=7cm]{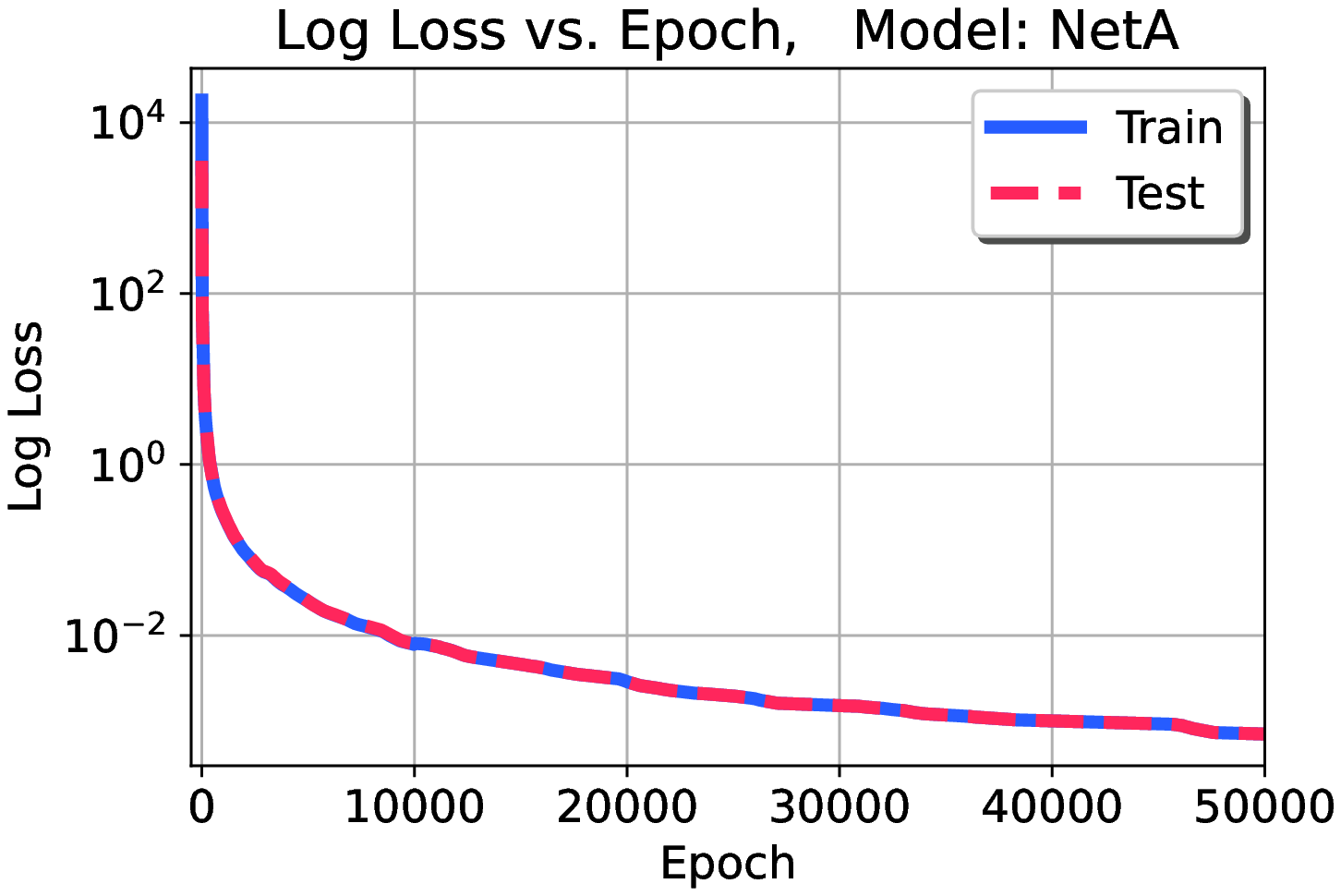}&\includegraphics[width=7cm]{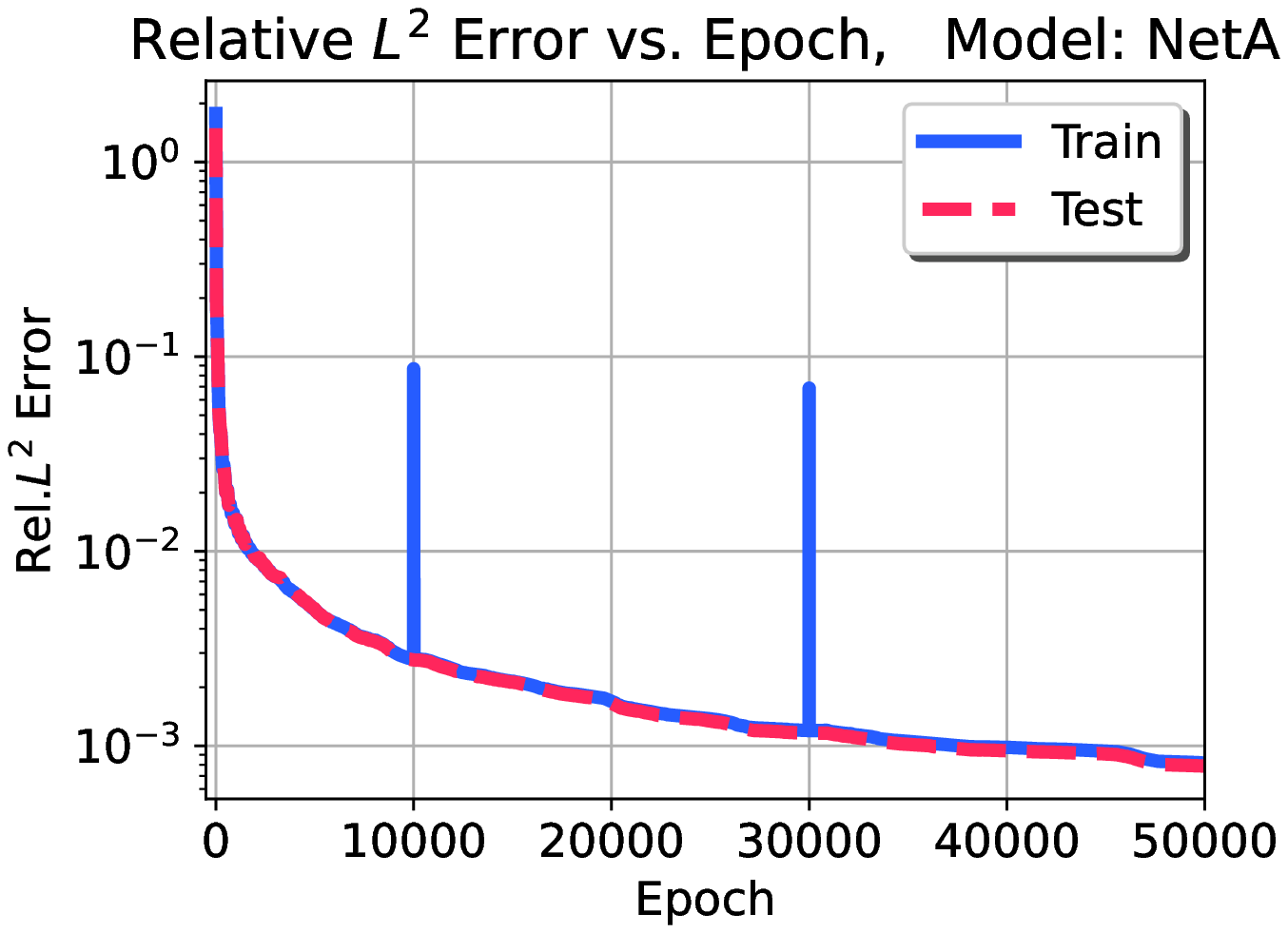}\\
     (a) training and test loss & (b) relative $L^2$ error\\
     \includegraphics[width=7cm]{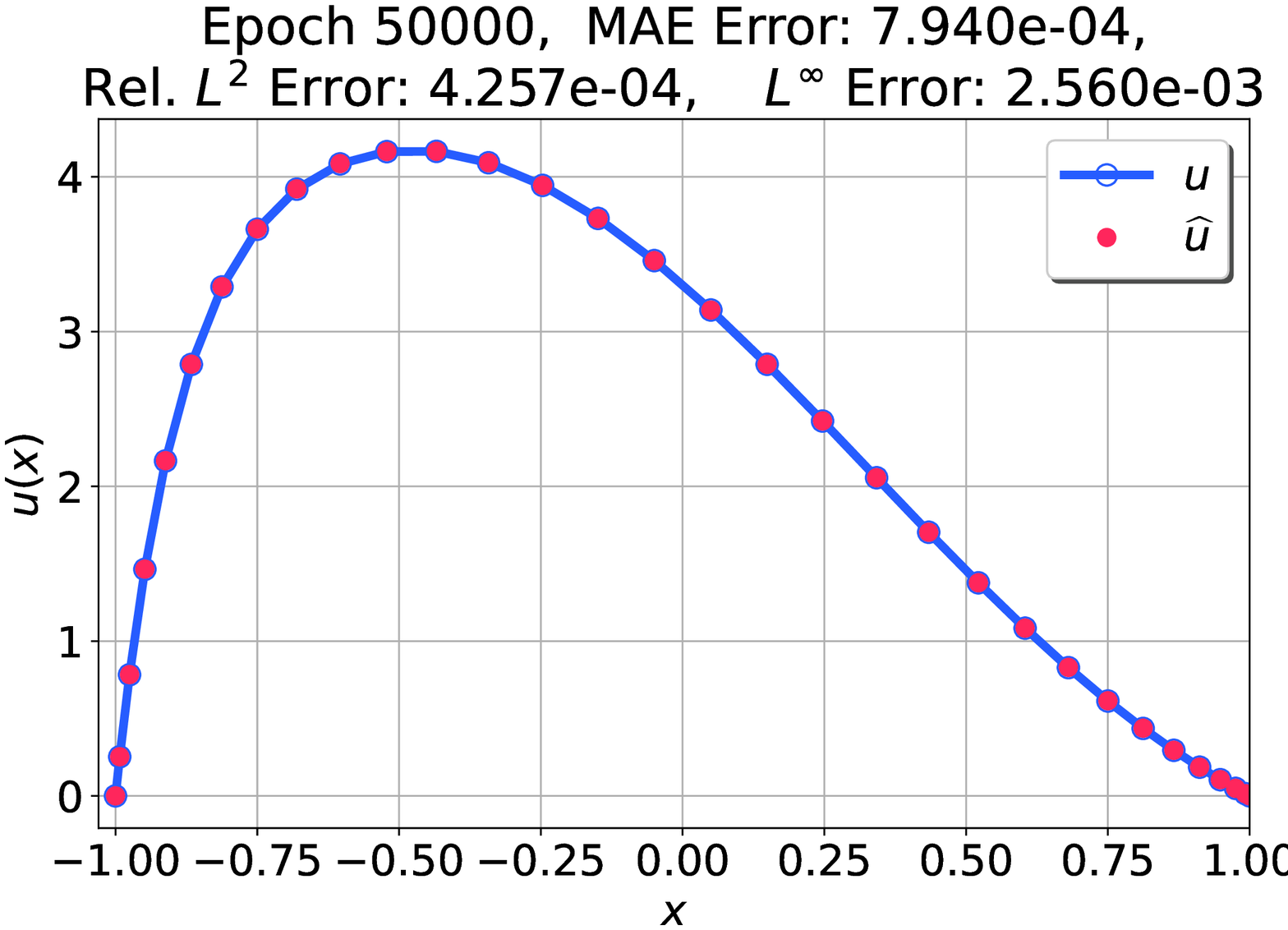}&\includegraphics[width=7cm]{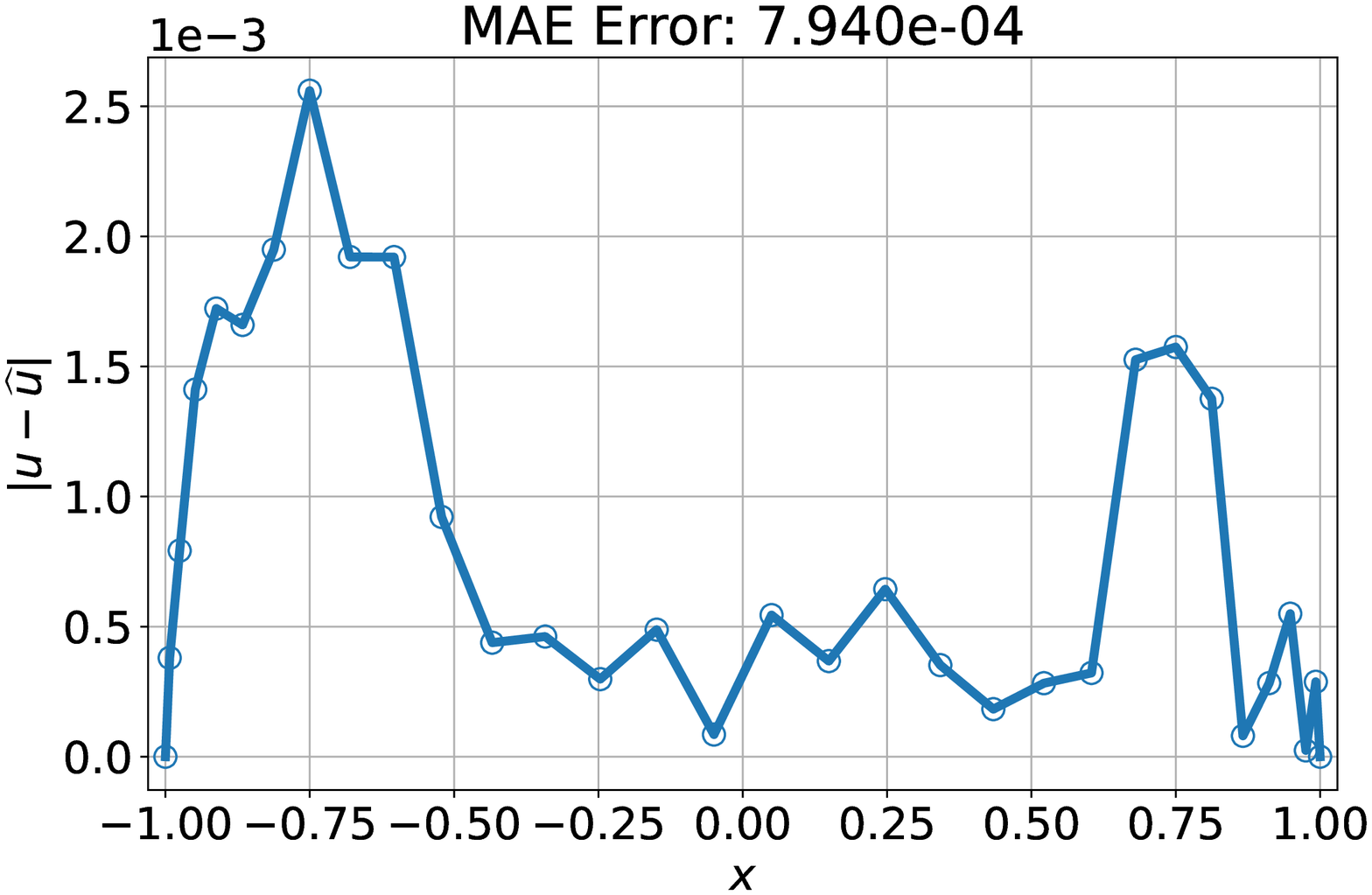}\\
(c) numerical solution $u$ &(d) point-wise error between $u$ and $\widehat{u}$\\
 and predicted solution $\widehat{u}$&     
     \end{tabular}
\caption{Numerical experiments of the CDE \eqref{CDE} with $\epsilon=10^{-1}$ are demonstrated. 
In panel (a), training and test loss curves are plotted on a semi-log scale up to $50,000$ epochs. 
Panel (b) shows the mean relative $L^2$ error stated in \eqref{rel_l2} between the true solution and predicted solution corresponding to the input.
Panel (c) shows that the predicted solution on out-of-sample input data is close to the corresponding exact solution with MAE: $7.940e-04$, relative $L^2$: $4.257e-04$, and $L^\infty$: $2.560e-03$. In panel (d), the point-wise error is plotted.}\label{CDE_fig}
\end{figure}

As shown in Figure \ref{CDE_fig} (a), the loss values \eqref{weak_loss} {for training and testing} decrease against the epoch.
%Our neural network can minimize generalization error.
The predicted solution is close to the corresponding exact solution on the order of $10^{-4}$ in relative $L^2$ errors. 
In Figure \ref{CDE_fig} (c), the predicted solution on out-of-sample data is close to the true solution with MAE: $7.940e-04$, relative $L^2$: $4.257e-04$, and $L^\infty$: $2.560e-03$.

\subsection{Helmholtz Equation with Neumann boundary condition}\label{HE_sect}
We turn our attention into the Helmholtz Equation with the Neumann boundary condition,
\begin{align}\label{HE}
\begin{split}
&u_{xx}+k_u u_x=f(x),\quad x\in (-1,1)=:\Omega\\
&u_x(-1)=u_x(1) = 0,
\end{split}
\end{align} 
where $k_u$ is a constant. 
{In order to impose the homogeneous Neumann boundary condition, we consider the compact form of the basis} as 
\begin{align}\label{basis_neumann}
\phi_k(x)=L_k(x)-\frac{k(k+1)}{(k+2)(k+3)} L_{k+2}(x),\quad x\in [-1,1].
\end{align}
Since the corresponding weak formulation to \eqref{HE} is defined by
\begin{align}\label{HE_weak}
-\int_{-1}^1 (\widehat{u}_i)_x\phi_xdx+\int_{-1}^1  k_u (\widehat{u}_i)_x\phi dx-\int_{-1}^1 = f_i\phi_j dx,
\end{align}
the loss function is written as
\begin{align}
loss=\sum_{i=1}^{P}\sum_{j=0}^{N-2}\left|\int_{-1}^1- (\widehat{u}_i)_x(\phi_j)_x+ k_u (\widehat{u}_i)_x\phi_j- f_i\phi_j dx\right|^2,
\end{align}
where $P$ is the number of inputs.
The architecture of the neural network is the same as that described in Section \ref{CDE_sect}.

\begin{figure}[t]
\begin{tabular}{cc}
     \includegraphics[width=7cm]{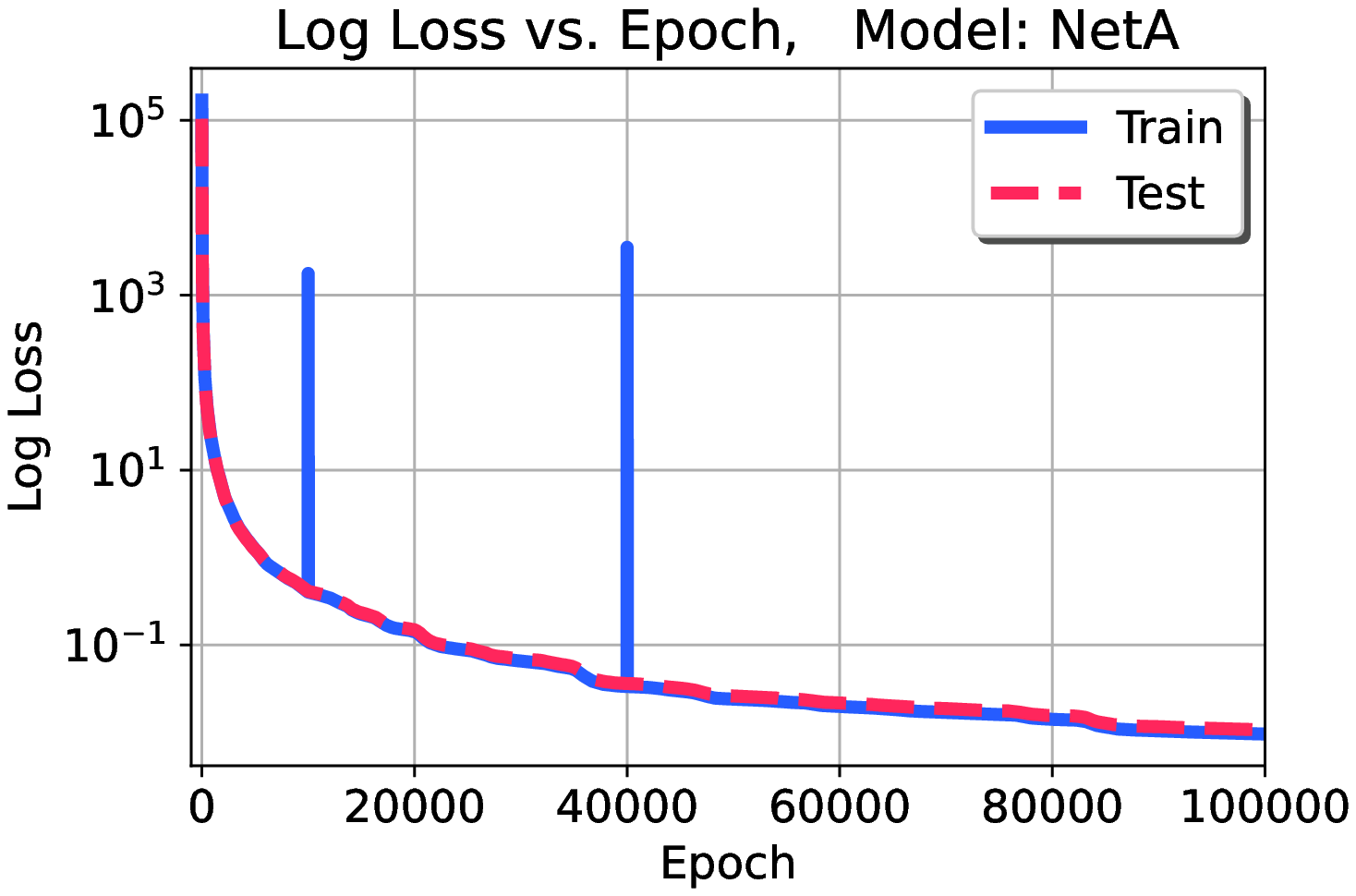}&\includegraphics[width=7cm]{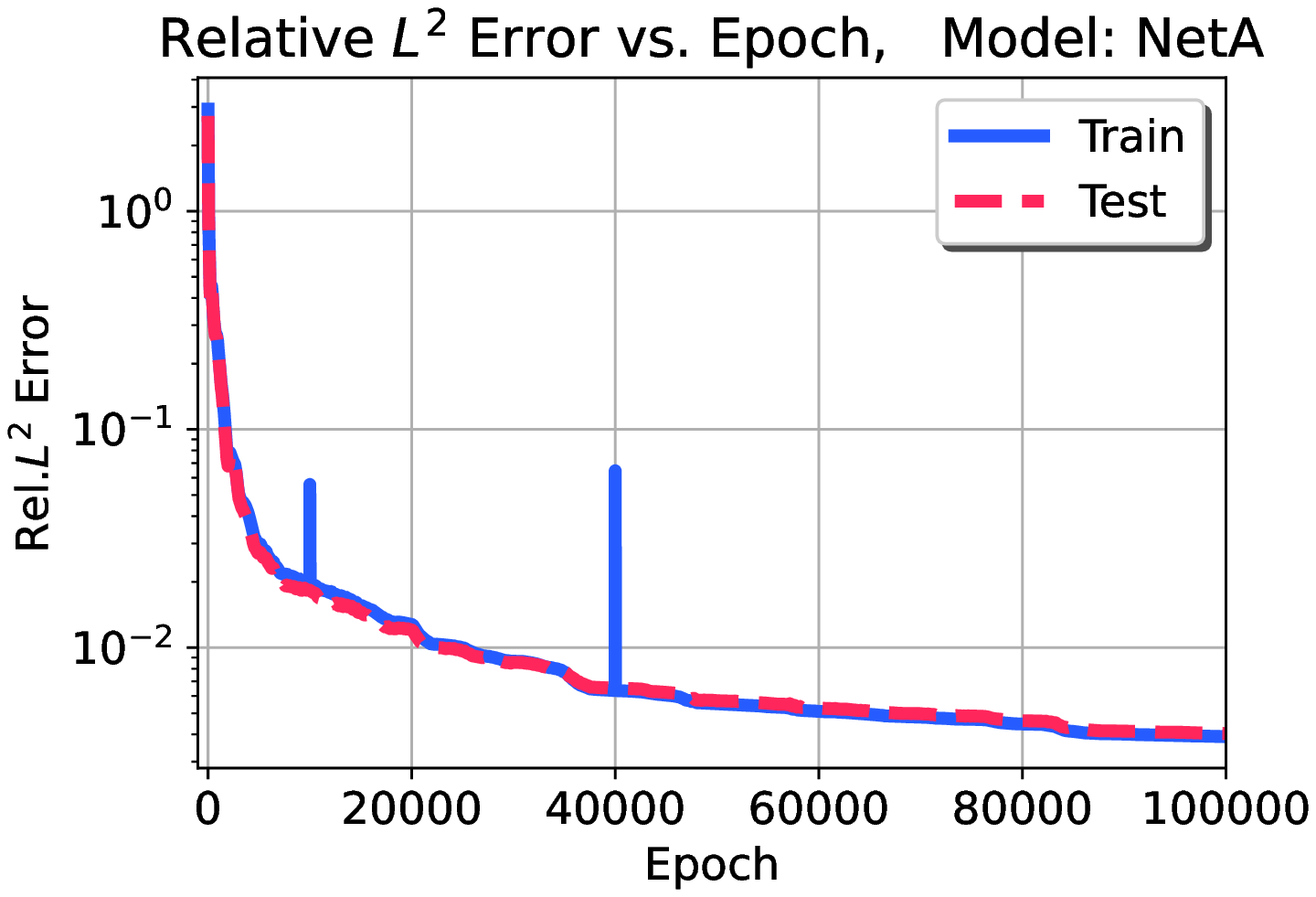}\\
     (a) training and test loss & (b) relative $L^2$ error\\
     \includegraphics[width=7cm]{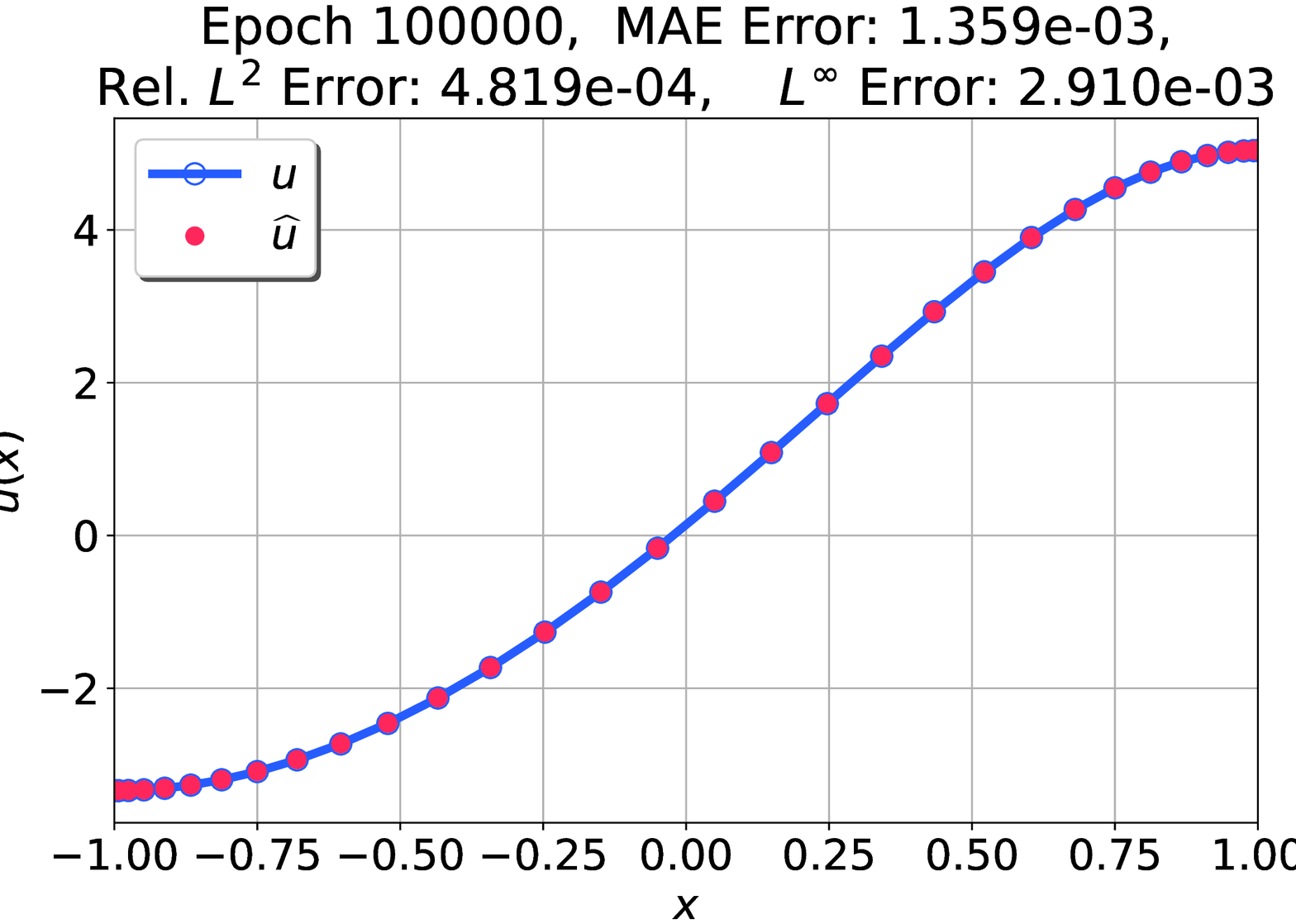}&\includegraphics[width=7cm]{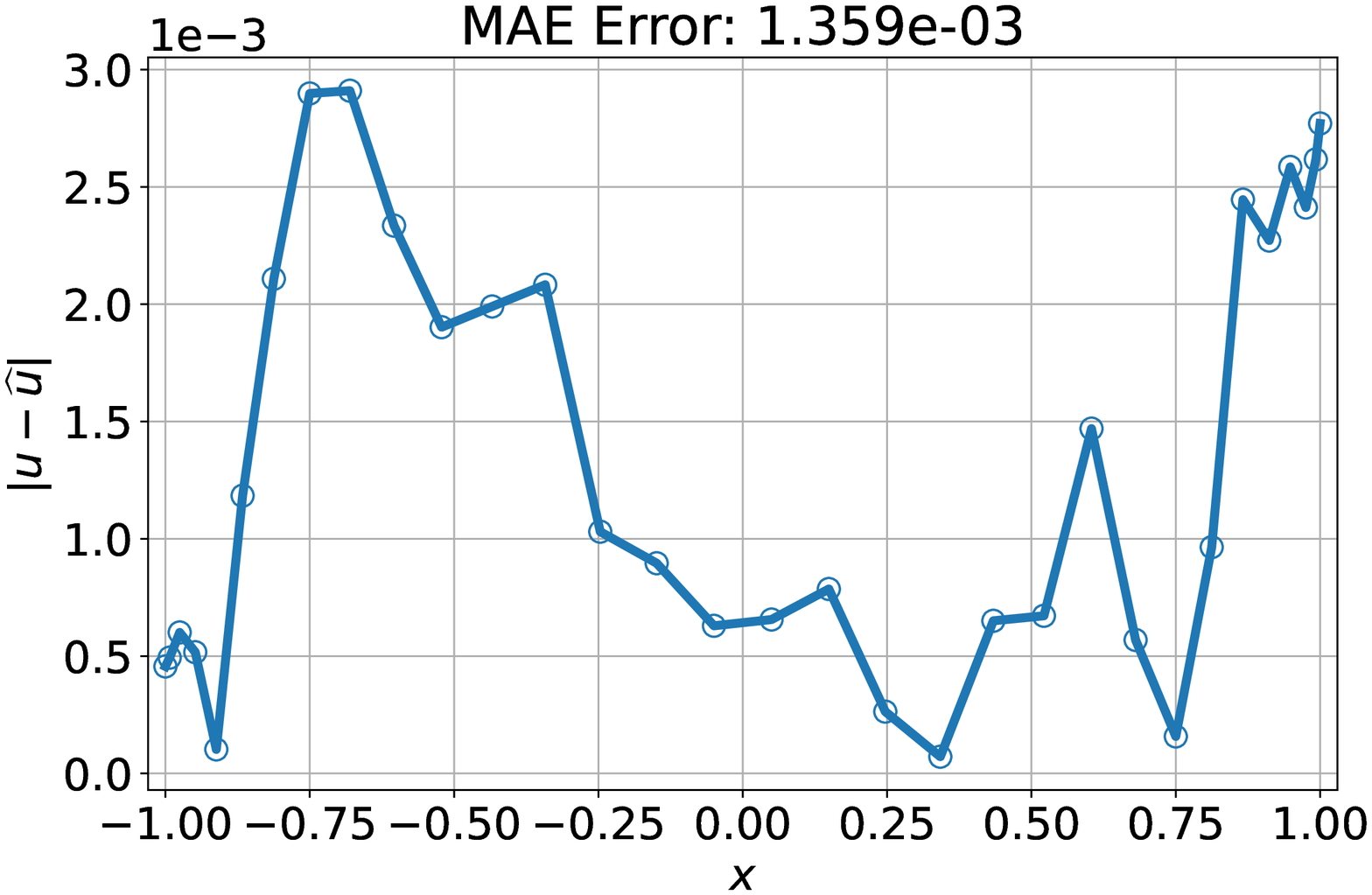}\\
     (c) numerical solution $u$ &(d) point-wise error between $u$ and $\widehat{u}$\\
 and predicted solution $\widehat{u}$&
     \end{tabular}
\caption{Numerical experiments of the Helmholz equation \eqref{HE} with $k_u = 3.5$ are displayed. 
In panel (a), training and test loss curves are plotted on a semi-log scale up to $10^5$ epochs. 
Panel (b) shows the mean relative $L^2$ error stated in \eqref{rel_l2} between the true solution and predicted solution corresponding to the input.
Panel (c) shows that the predicted solution on out-of-sample input data is close to the corresponding exact solution with MAE: $1.359e-03$, relative $L^2$: $4.819e-04$, and $L^\infty$: $2.910e-03$. In panel (d), the point-wise error is plotted.}\label{HE_fig}
\end{figure}

Figure \ref{HE_fig} (a) shows {the losses values of the inputs for training
and testing} against the epoch. 
Figure \ref{HE_fig} (c) indicates that the predicted solution on the out-of-sample data is close to the corresponding the exact solution, with MAE:$4.248e-05$, relative $L^2$: $4.819e-04$, and $L^\infty$: $2.910e-03$.
In fact, the numerical solution predicted by ULGNet exactly assigns the homogeneous Neumann boundary condition, the numerical solutions attain good accuracy.

\subsection{Burgers' equation}
In the section, we apply the proposed method to the nonlinear Burgers' equations
\begin{align}\label{BE}
\begin{split}
&-u_{xx}+uu_x=f(x),\\
&u(-1)=u(1) = 0.
\end{split}
\end{align}
The corresponding weak formulation to \eqref{BE} is defined by
\begin{align}\label{BE_weak}
\int_{-1}^1 {u}_x\phi_x-\int_{-1}^1 \frac{{u}^2}{2}\phi_x-\int_{-1}^1 f\phi dx=0,
\end{align} 
where $\phi_j$'s are described in \eqref{basis_dirichlet}.
The loss function is written as
\begin{align}
loss=\sum_{i=1}^P\sum_{j=0}^{N-2}\left|\int_{-1}^1 (\widehat{u}_i)_x(\phi_j)_x-\frac{(\widehat{u}_i)^2}{2}(\phi_j)_x- f_i\phi_jdx\right|^2,
\end{align}
where $\phi_j$'s are as in \eqref{basis_dirichlet} and $P$ is the number of inputs.
In order to deal with the nonlinearity, we use the $Swish$ activation function, as
\begin{align}
Swish(x)=\frac{x}{1+\exp(-x)}.
\end{align}
The architecture of the neural network is as described in Section \ref{HE_sect}.

\begin{figure}[t]
\begin{tabular}{cc}
     \includegraphics[width=7cm]{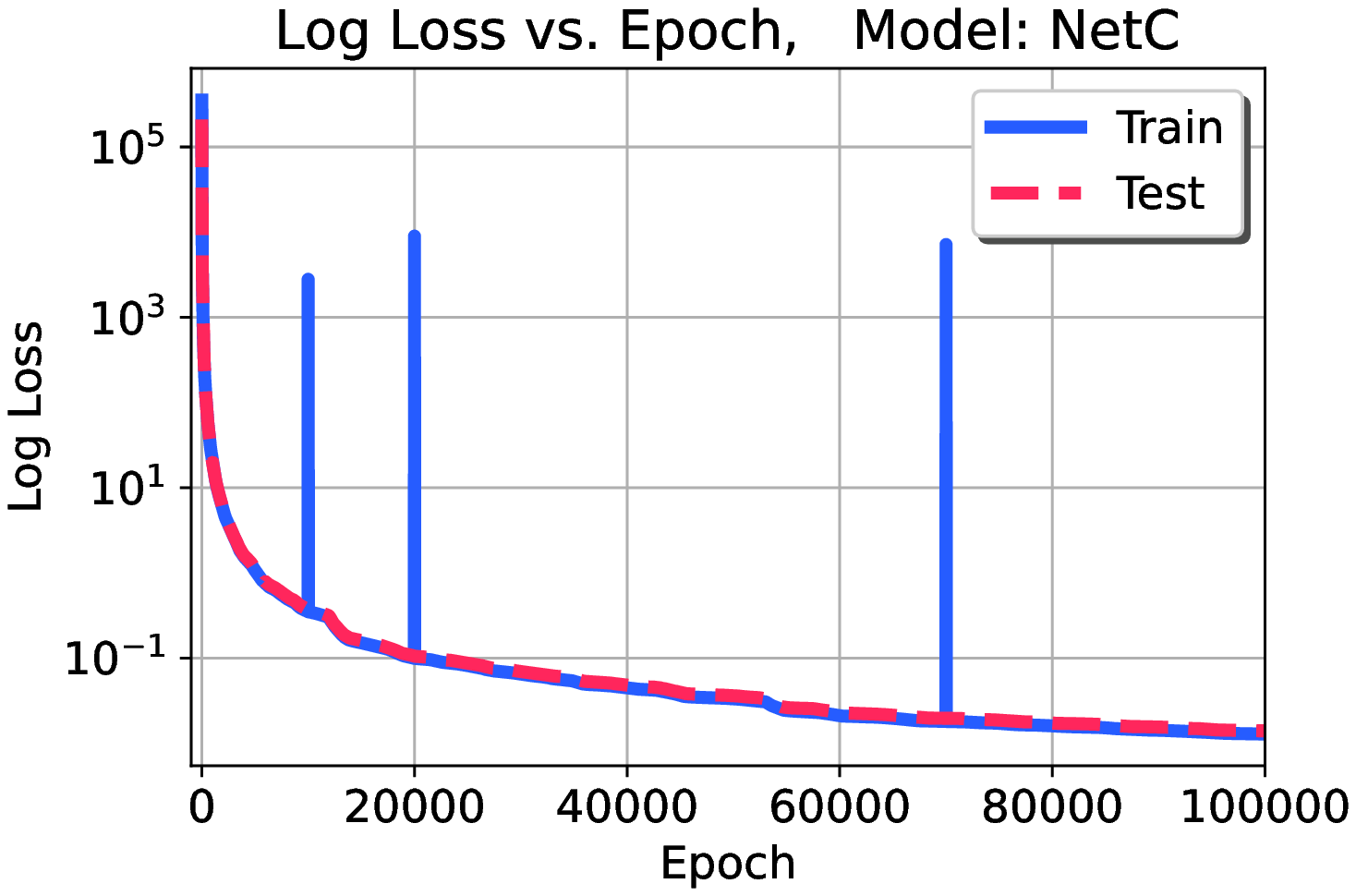}&\includegraphics[width=7cm]{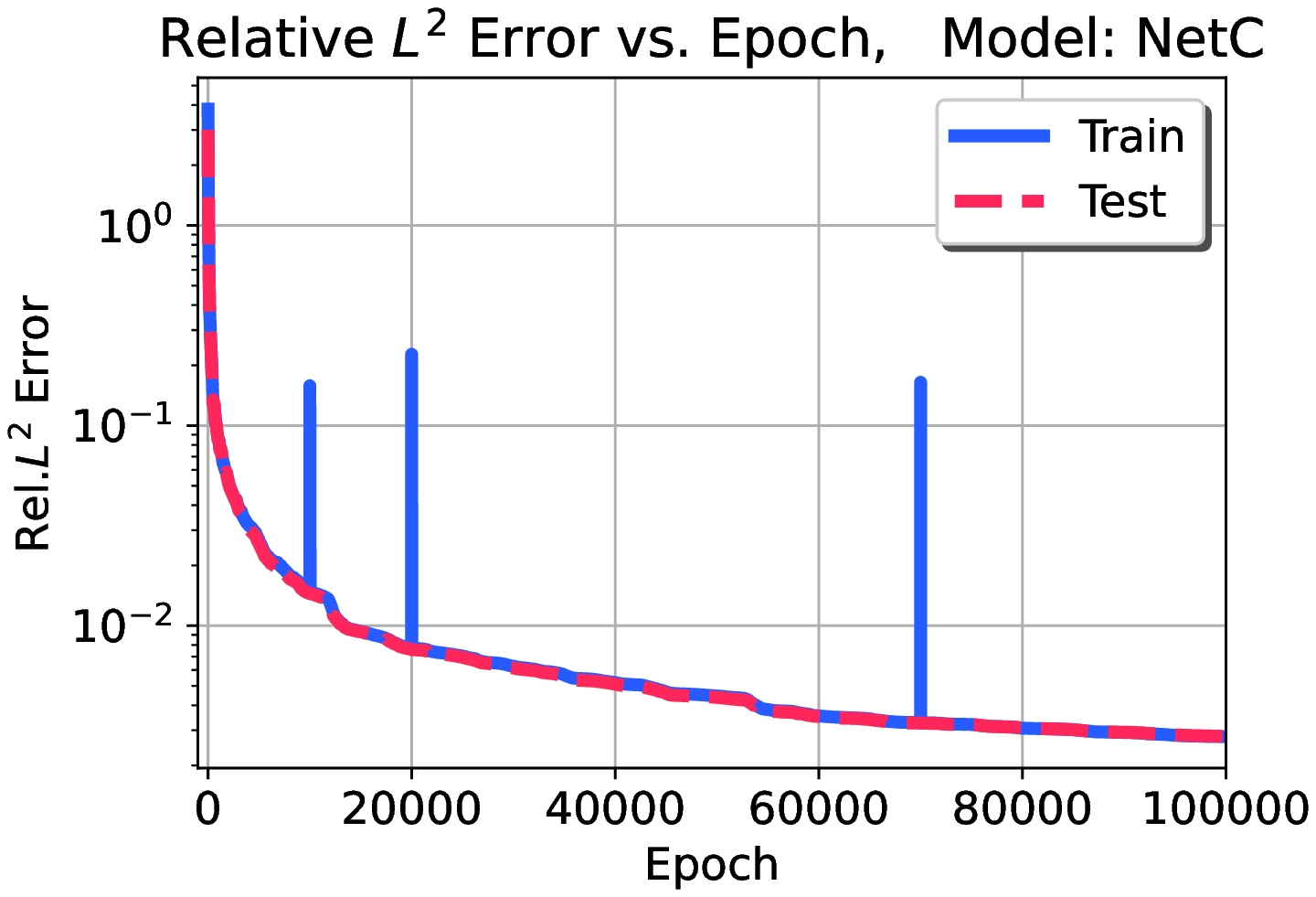}\\
     (a)  training and test loss  & (b) relative $L^2$ error\\
     \includegraphics[width=7cm]{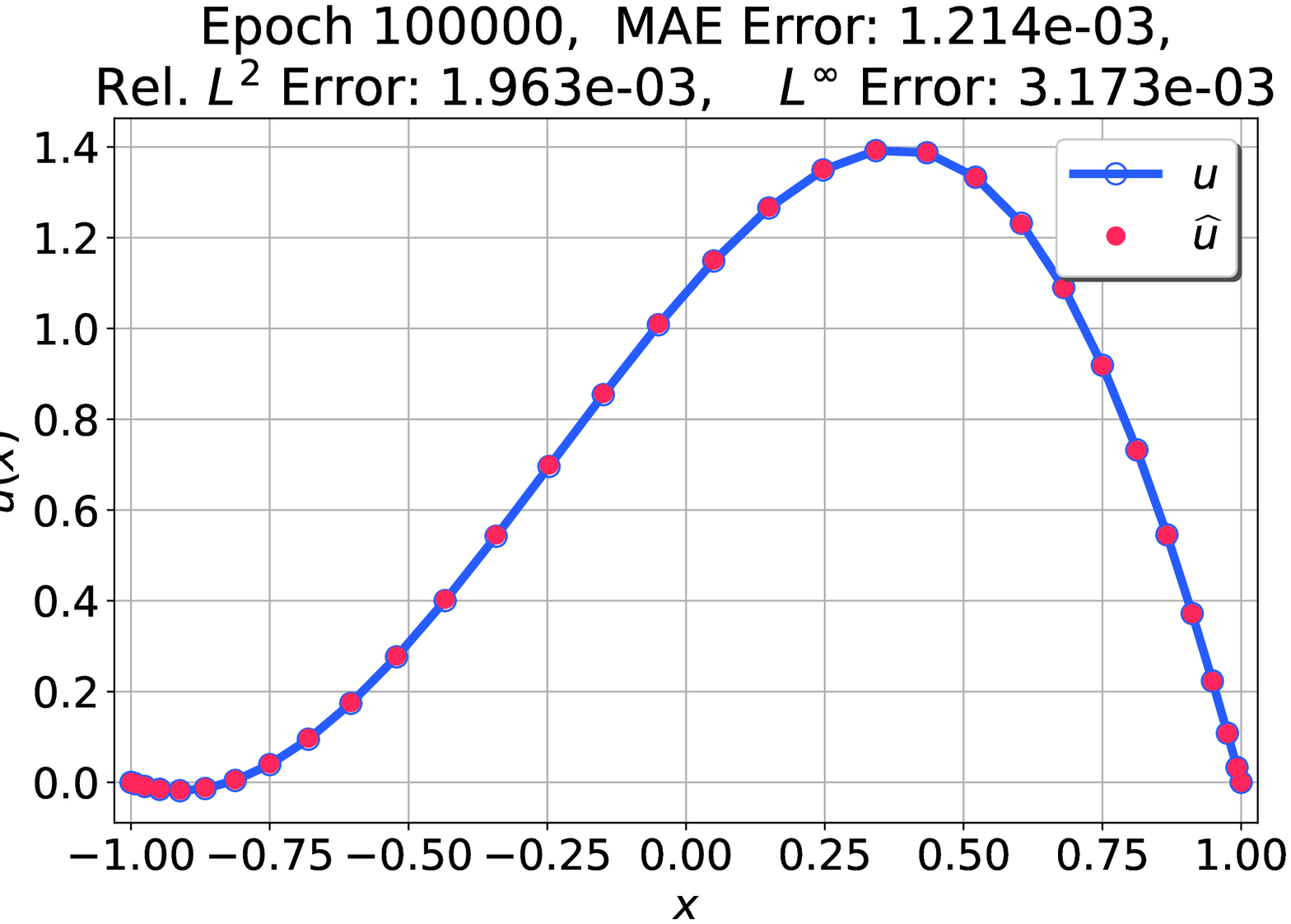}&\includegraphics[width=7cm]{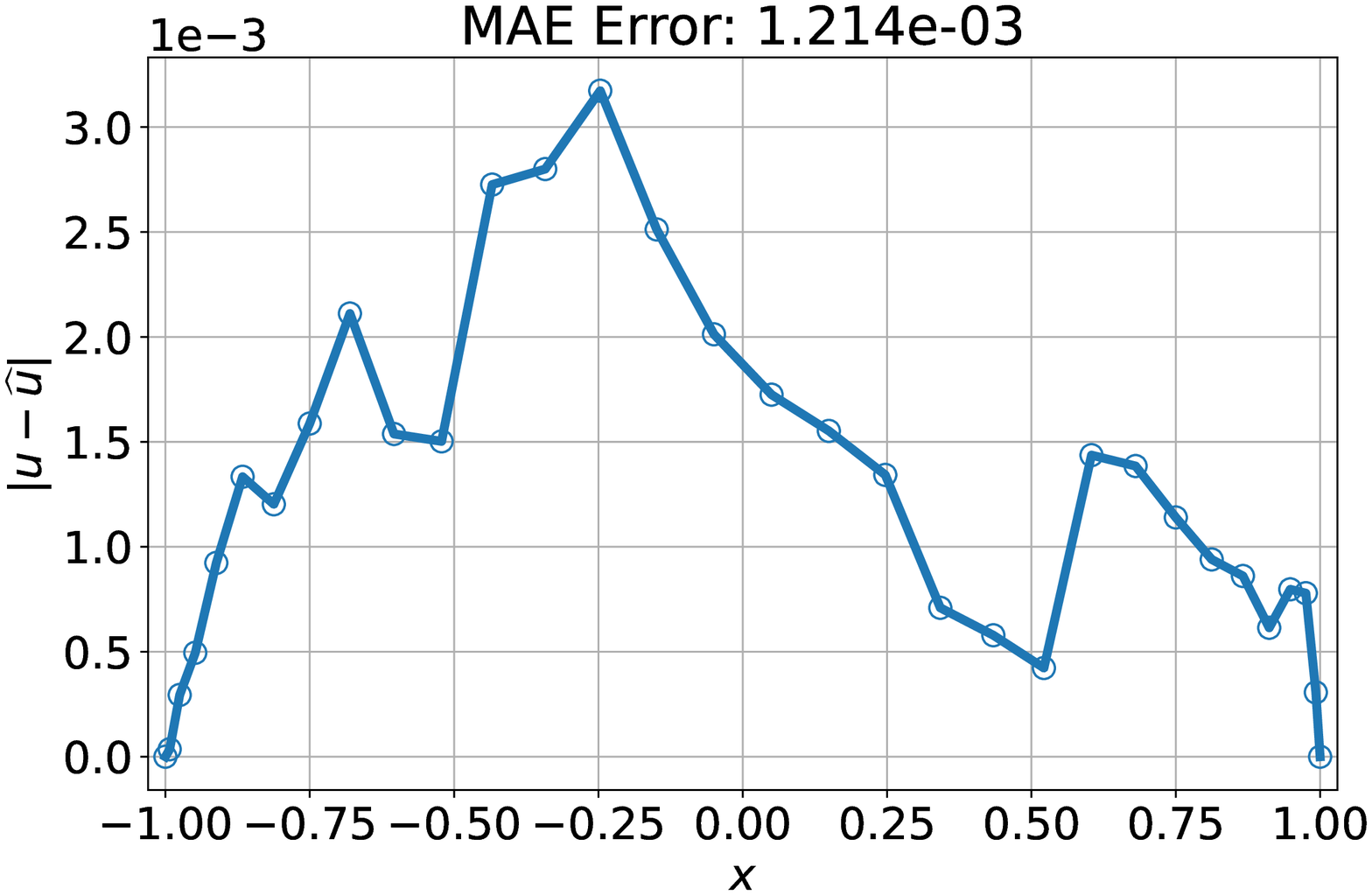}\\
     (c) numerical solution $u$ &(d) point-wise error between $u$ and $\widehat{u}$\\
 and predicted solution $\widehat{u}$&
     \end{tabular}
\caption{Numerical experiments of the Burgers' equation \eqref{BE} are displayed. 
In panel (a), training and test loss curves are plotted on a semi-log scale up to $10^5$ epochs. 
Panel (b) shows the mean relative $L^2$ error stated in \eqref{rel_l2} between the true solution and predicted solution corresponding to the input.
Panel (c) shows that the predicted solution on out-of-sample input data is close to the corresponding exact solution with MAE: $1.214e-03$, relative $L^2$: $1.963e-03$, and $L^\infty$: $3.173e-03$. In panel (d), the point-wise error is plotted.}\label{BE_fig}
\end{figure}

In Figure \ref{BE_fig} (a), the losses for training
and testing decrease against the epoch. 
Figure \ref{BE_fig} (c) shows that the predicted solution on out-of-sample data is close to the corresponding exact solution with MAE:$1.214e-03$, relative $L^2$: $1.963e-03$, and $L^\infty$: $3.173e-03$.

\begin{figure}[t]
\begin{tabular}{cc}
     \includegraphics[width=7cm]{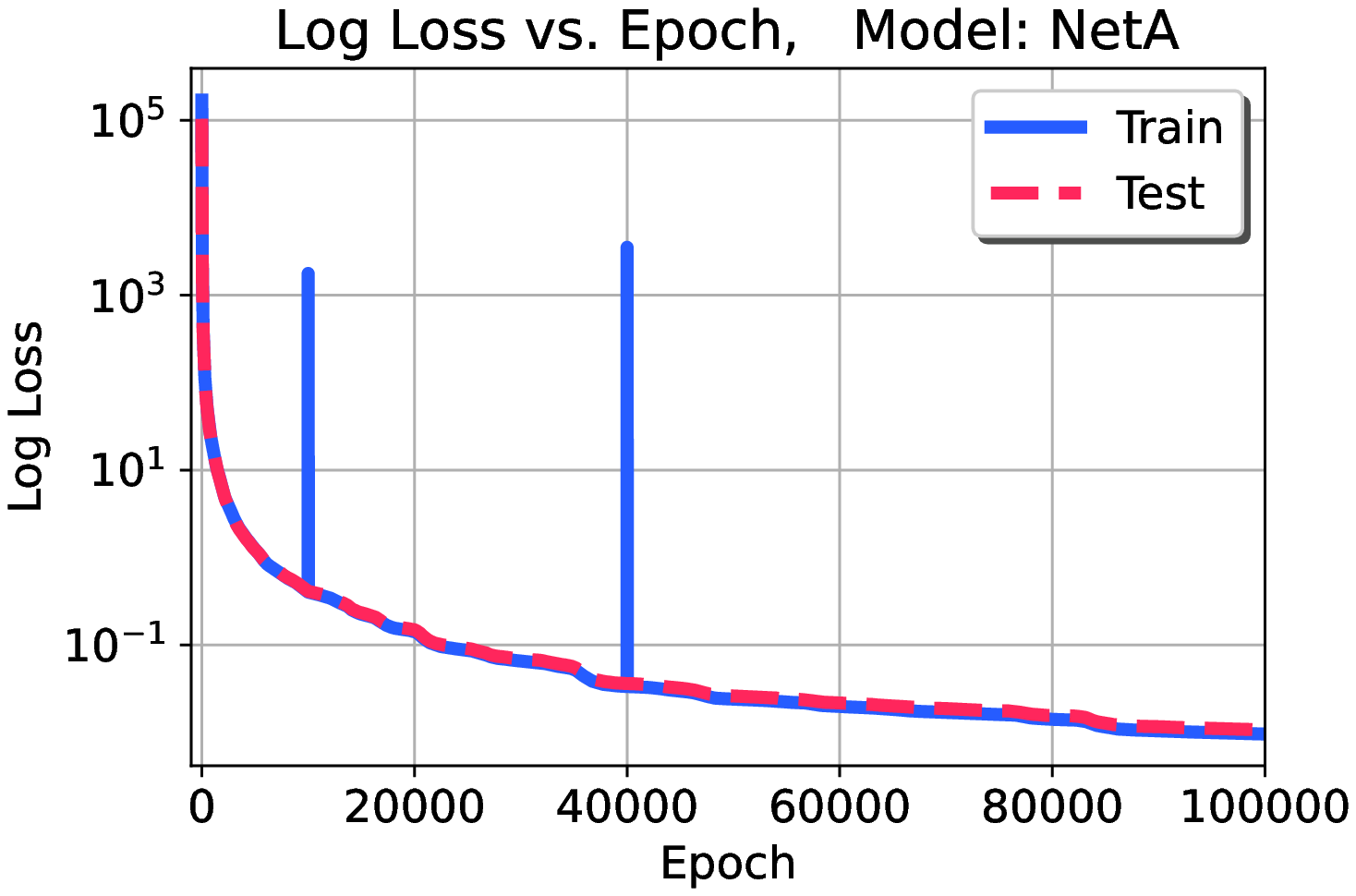}&\includegraphics[width=7cm]{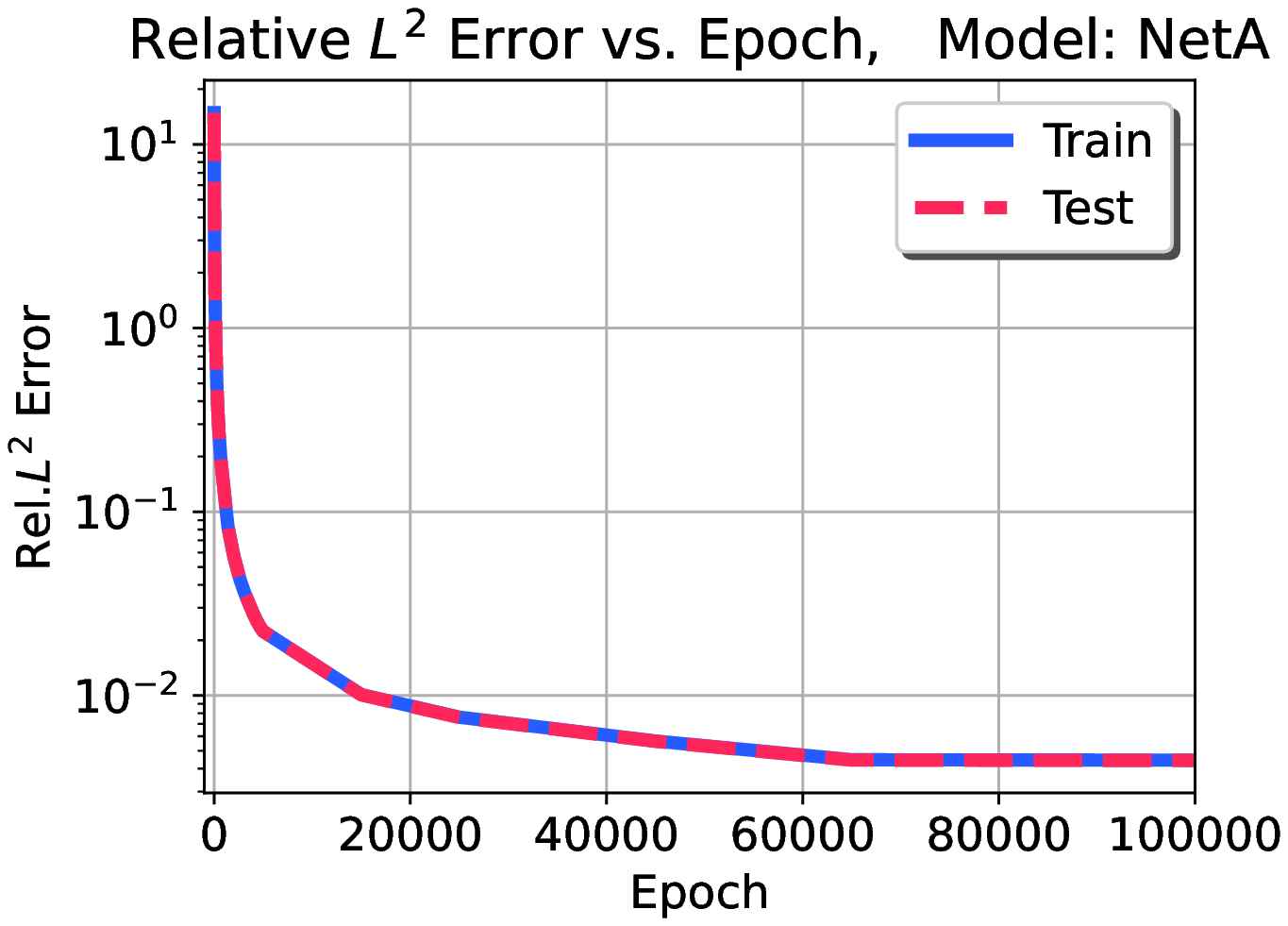}\\
     (a)  training and test loss  & (b) relative $L^2$ error
        \end{tabular}     
     \includegraphics[width=15cm]{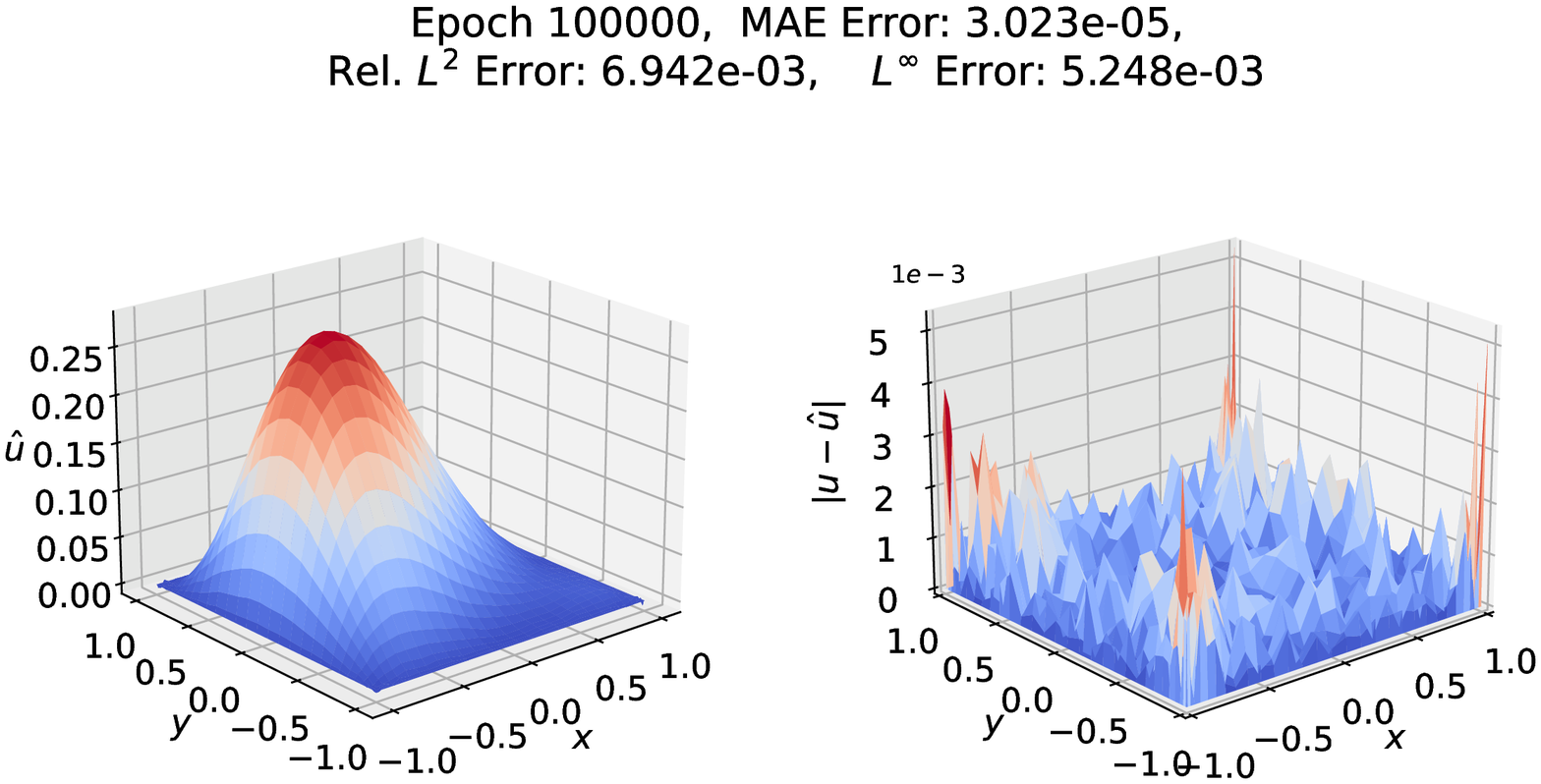}\\
     (c) predicted solution $\widehat{u}$ (left) and point-wise error between $u$ and $\widehat{u}$ (right).
\caption{Numerical experiments of the 2D CDE \eqref{2D_CDE} with $\epsilon=10^{-1}$ and ${\bf{v}}=(-1,0)$ are displayed. 
In panel (a), training and test loss curves are plotted on a semi-log scale up to $10^5$ epochs. 
Panel (b) shows the mean relative $L^2$ error stated in \eqref{rel_l2} between the true solution and predicted solution corresponding to the input.
Panel (c) shows that the predicted solution on out-of-sample input data is close to the corresponding exact solution with MAE: $1.214e-03$, relative $L^2$: $1.963e-03$, and $L^\infty$: $3.173e-03$. The point-wise error is plotted in the right panel.}\label{2D_CDE_fig}
\end{figure}

\subsection{Two-dimensional convection-diffusion equation}
In this section, we consider the two dimensional convection-diffusion equation,
\begin{align}\label{2D_CDE}
\begin{split}
-\epsilon \Delta u+{\bf{v}}\cdot\nabla u&=f(x),\quad x\in[-1,1]\times[-1,1]=:\Omega\subset\mathbb{R}^2\\
u&=0,\qquad~ x\in\partial\Omega.
\end{split}
\end{align} 
To construct the two dimensional basis functions with Dirichlet BC, we naturally employ $\phi_{ij}(x,y) = \phi^x_i(x)\phi^y_j(y)$ where $\phi^x_i$ and $\phi^y_j$ are basis in $x$ and $y$ direction as in \eqref{basis_dirichlet}, respectively.
The corresponding weak formulation in \eqref{2D_CDE} is defined by
\begin{align}\label{2D_CDE_weak}
\int_{\Omega} \epsilon \nabla u\cdot\nabla\phi_{ij} dx+\int_{\Omega} {\bf{v}}\cdot\nabla u \phi_{ij} dx =\int_{\Omega} f\phi_{ij} dx.
\end{align}
We then obtain the loss function as 
\begin{align}\label{2D_CDE_loss}
loss=\sum_{p=1}^{P}\sum_{i=0}^{N-2}\sum_{j=0}^{M-2}\left|\int_{\Omega} \epsilon \nabla\widehat{u}_p\cdot\nabla\phi_{ij}+ {\bf{v}}\cdot\nabla \widehat{u}_p \phi_{ij}- f_p\phi_{ij}dx\right|^2
\end{align}
where $P$ is the number of inputs.
We generate forcing functions as input data
\begin{align}\label{EF2D}
f_p(x,y)=h_{1p}\sin(\pi (m_{1p}x+m_{2p}y))+h_{2p}\cos(\pi (m_{3p}x+m_{4p}y)),
\end{align}
where $h_{ip}$ for $i=1,2$ and $m_{jp}$ for $1 \leq j \leq 4$ are drawn from a uniform distribution on $[1,2]$ and $[0,1]$, respectively and $p=1,2,\cdots,P$. For numerical experiments, we set $\epsilon=0.1$ and ${\bf{v}}=(-1,0)$.

As shown in Figure \ref{2D_CDE_fig} (a), the losses for the training and test decrease against the epoch. 
Figure \ref{2D_CDE_fig} (c) shows that the predicted solution on out-of-sample is close to the corresponding exact solution with MAE: $3.023e-05$, relative $L^2$: $6.942e-03$, and $L^\infty$: $5.248e-03$. 

\begin{figure}[t]
\begin{tabular}{cc}
     \includegraphics[width=7cm]{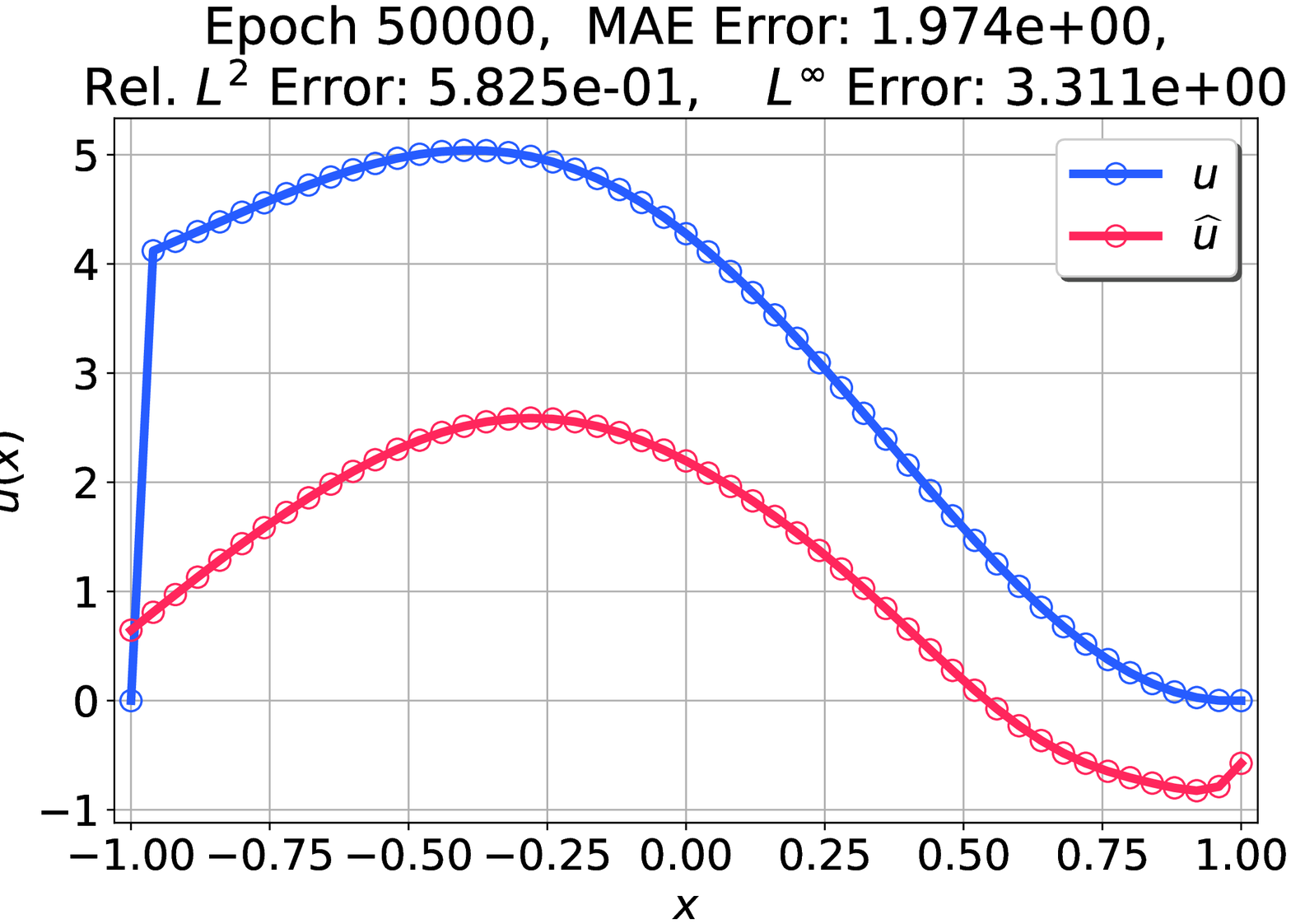}&\includegraphics[width=7cm]{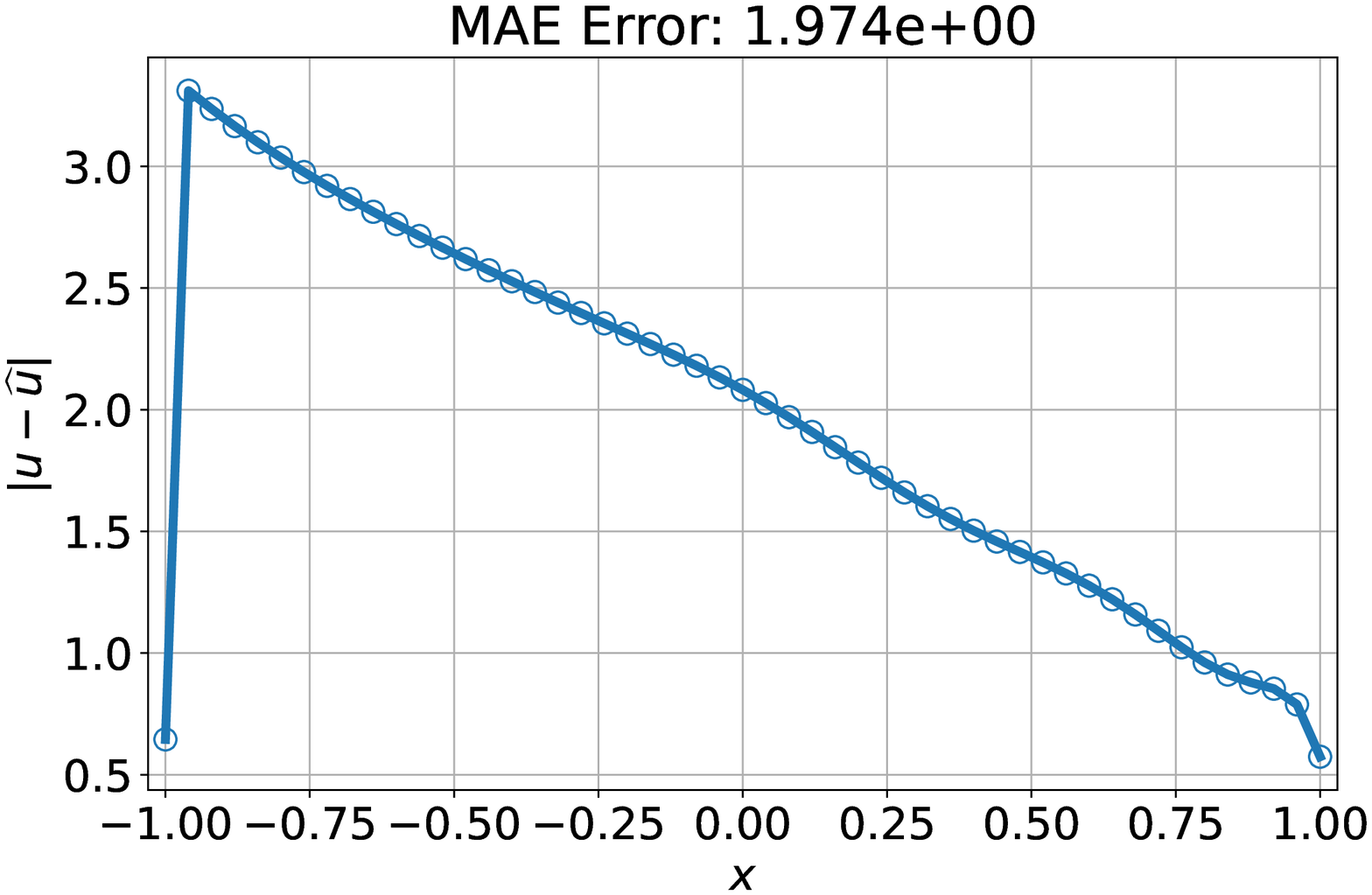}\\
     \end{tabular}\\
     {(a) Numerical experiment using the PINN method for \eqref{CDEB} with $\epsilon = 10^{-6}$.
     Left: reference solution $u$ and predicted solution $\widehat{u}$ predicted by the PINN.
     Right: point-wise error between $u$ and $\widehat{u}$. }
      \begin{tabular}{cc}
     \includegraphics[width=7cm]{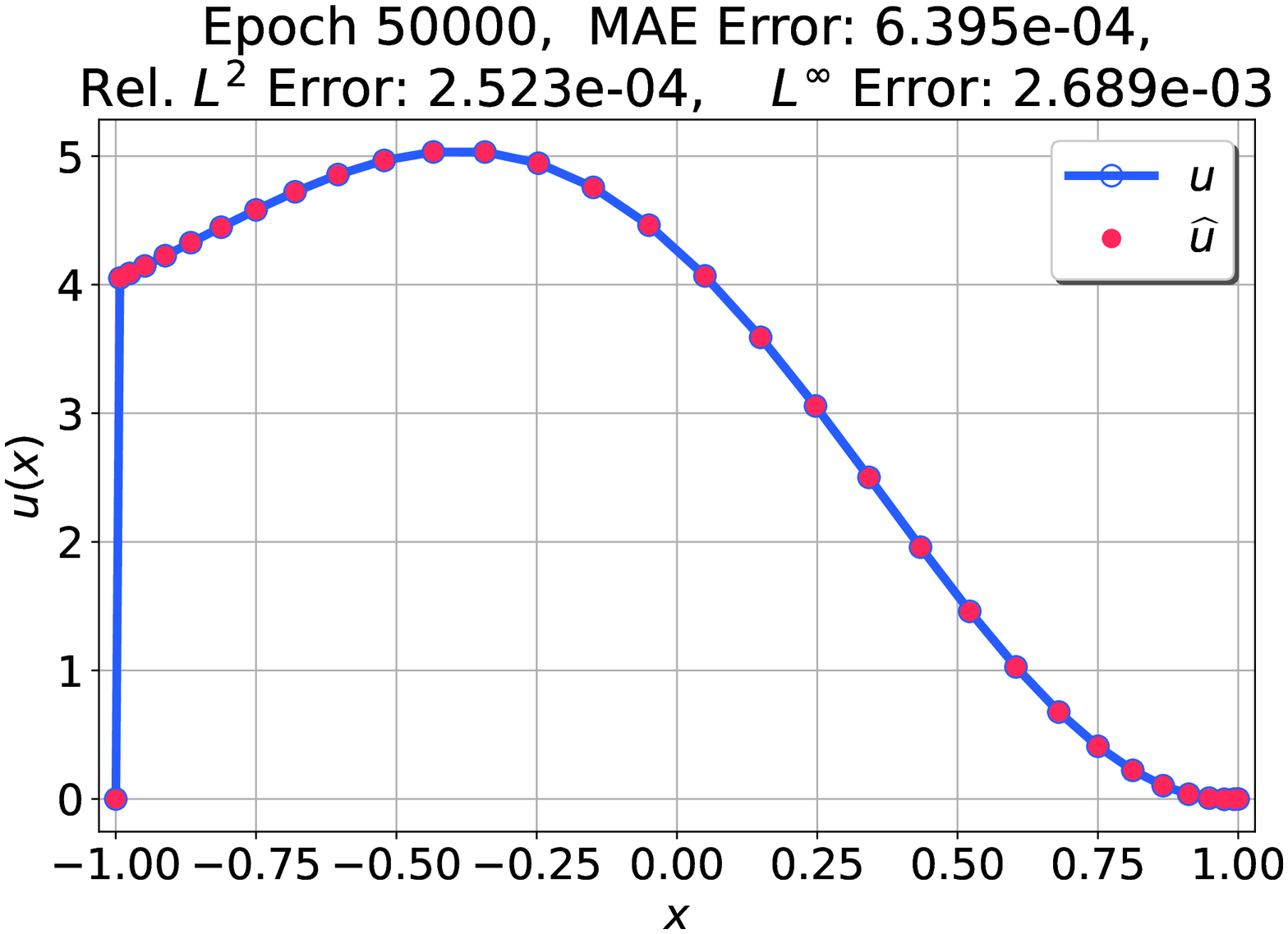}&\includegraphics[width=7cm]{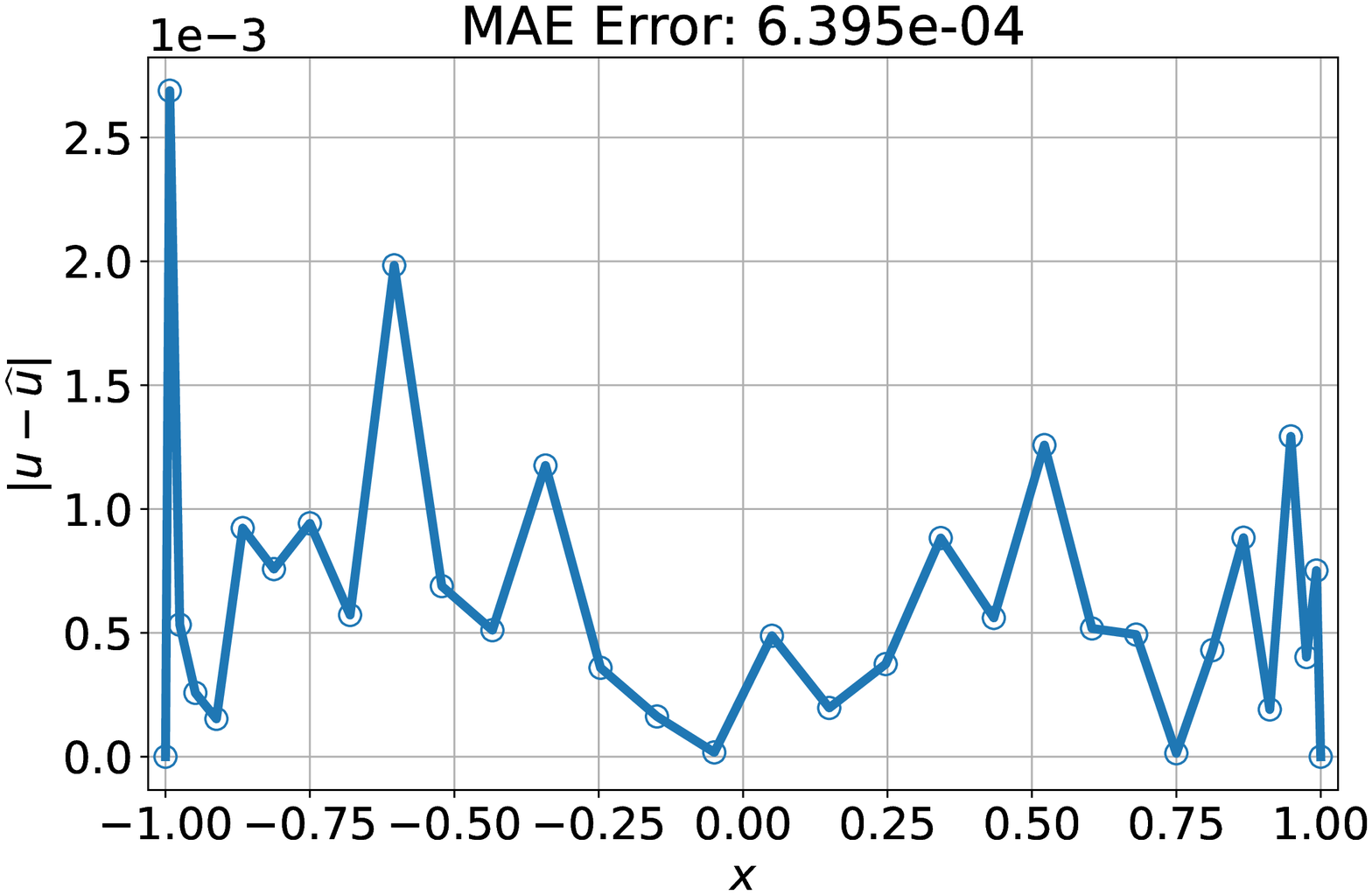}\\     
     \end{tabular}\\    
      {(b) Numerical experiment using ULGNet (ours) for \eqref{CDEB} with $\epsilon = 10^{-6}$.
     Left: reference solution $u$ and predicted solution $\widehat{u}$ predicted by the ULGNet.
     Right: point-wise error between $u$ and $\widehat{u}$. }
\caption{{Comparison between PINN and ULGNet (Ours) for the CDE \eqref{CDEB} with $\epsilon=10^{-6}$ is displayed. 
While the PINN solution in panel (a) produces a large numerical error, the ULGNet solution in panel (b) shows an accurate numerical solution.}}\label{CDEB_fig}
\end{figure}

\section{Numerical experiments: singular perturbation problem}
\label{sec:4}
In numerical analysis, convection-dominated singularly perturbed problems is difficult to solve without special treatment since a small diffusive parameter produces a sharp transition inside thin layers.
As in scientific computing, ML approaches suffer from the boundary layer issue since neural networks rely on the
smooth prior.
In this regard, we develop basis enriched method for our ULGNet to resolve the boundary layer issue.
Our numerical experiments manifest that our algorithm is effective on finding an accurate numerical solution of singularly perturbed convection-diffusion equations which exhibit strong boundary layer behavior.

%As aforementioned, the performance of PINN scheme suffers for stiff PDEs \cite{wang2022and}. However, our scheme can resolve the stiffness by enriching a set of bases with an additional basis function. 

\subsection{Convection-diffusion equation with a boundary layer: one-dimensional case}\label{sect_CDEB}
We propose our enriched algorithm to the convection-diffusion equation in one dimension,
\begin{align}\label{CDEB}
\begin{split}
&-\epsilon u_{xx}-u_x=f,\\
&u(-1)=u(1)=0,
\end{split}
\end{align} 
with $\epsilon \ll 1$. 
In contrast to Section \ref{CDE_sect}, due to the structure of \eqref{CDEB}, the solution profile has a sharp transition near $x=-1$, which is referred to as a boundary layer. 
Capturing the stiff boundary layer is challenging in numerical analysis, but understanding the boundary layer phenomenon is important in many scientific applications. 
In order to handle the boundary layer as in \cite{hong2018enriched} and \cite{jung2005numerical}, the additional basis function can be used
\begin{align}\label{cor}
\phi_{cor}= \exp(-(1+x)/\epsilon)-\left(1- \frac{1 - \exp(-2/\epsilon)}{2}(x+1)\right)
\end{align}      
named a boundary layer corrector (for the specific derivation, see the Appendix). %In fact, the corrector function corresponds to the boundary layer profile within a thin layer.
The former exponential term in \eqref{cor} stands for a boundary layer profile. 
After adding the latter linear term to the former, \eqref{cor} satisfies the Dirichlet BC. 
We then enrich \eqref{inference} as
\begin{equation} \label{e:e_sum_form}
    \sum^{N-2}_{j=0}  \alpha_j \phi_j +\alpha_{N-1} \phi_{N-1},
\end{equation}
where $\phi_{N-1} = \phi_{cor}$ in \eqref{cor}.
We define the loss function with the $N-1$ base functions as
\begin{align}
loss=\sum_{i=1}^P\sum_{j=0}^{N-1}\left|\int_{-1}^1 \epsilon(\widehat{u}_i)_x(\phi_j)_x-(\widehat{u}_i)_x\phi_j-f_i\phi_j\right|^2.
\end{align}
For numerical experiments below, we set $\epsilon = 10^{-6}$. %{\color{blue}In addition, in order to enhance the performance, the experiments in this section are conducted with double precision floating format.}
Figure \ref{CDEB_fig} depicts that the predicted solution using the ULGNet on the out-of-sample input is close to the corresponding exact solution within MAE: $6.395e-04$, relative $L^2$: $2.523e-04$, and $L^\infty$: $2.689e-03$.  
For comparison, we provide an experiment using the PINN method in panel (a) of figure \ref{CDEB_fig}.
The numerical results manifest that the ULGNet can provide an accurate numerical solution to the convection-dominated singularly perturbed problem while the standard PINN method do not the sharp transition effectively.
% \begin{rem}
% Considering spectral method with LGL basis, the formulation for inner products between bases such as $(\phi_i,\phi_j)_{H^1_0}$, is well known in \cite{shen2011spectral}. In the case for $\phi_{30}$ \eqref{cor}, the inner product can be directly computed. The process is provided in the Appendix.    
% \end{rem}

% \begin{figure}[t]
% \begin{tabular}{cc}
%      \includegraphics[width=7cm]{figure/pinn_loss.eps}&\includegraphics[width=7cm]{figure/pinn_loss_u.eps}\\
%      (a) Losses & (b) Relative $L^2$ error\\
%      & between the training and test sets \\
%      \includegraphics[width=7cm]{figure/pinn_u_upre.eps}&\includegraphics[width=7cm]{figure/pinn_err.eps}\\
%      (c) Numerical solution $u$ &(d) Point-wise error between $u$ and $\widehat{u}$\\
%  and predicted solution $\widehat{u}$&     
%      \end{tabular}
% \caption{ {\color{blue} A reference experiment on the identical conditions to figure \ref{CDEB_fig} but with PINN. }    }\label{Pinn1d_fig}
% \end{figure}
\begin{figure}[t]

     \includegraphics[width=15cm]{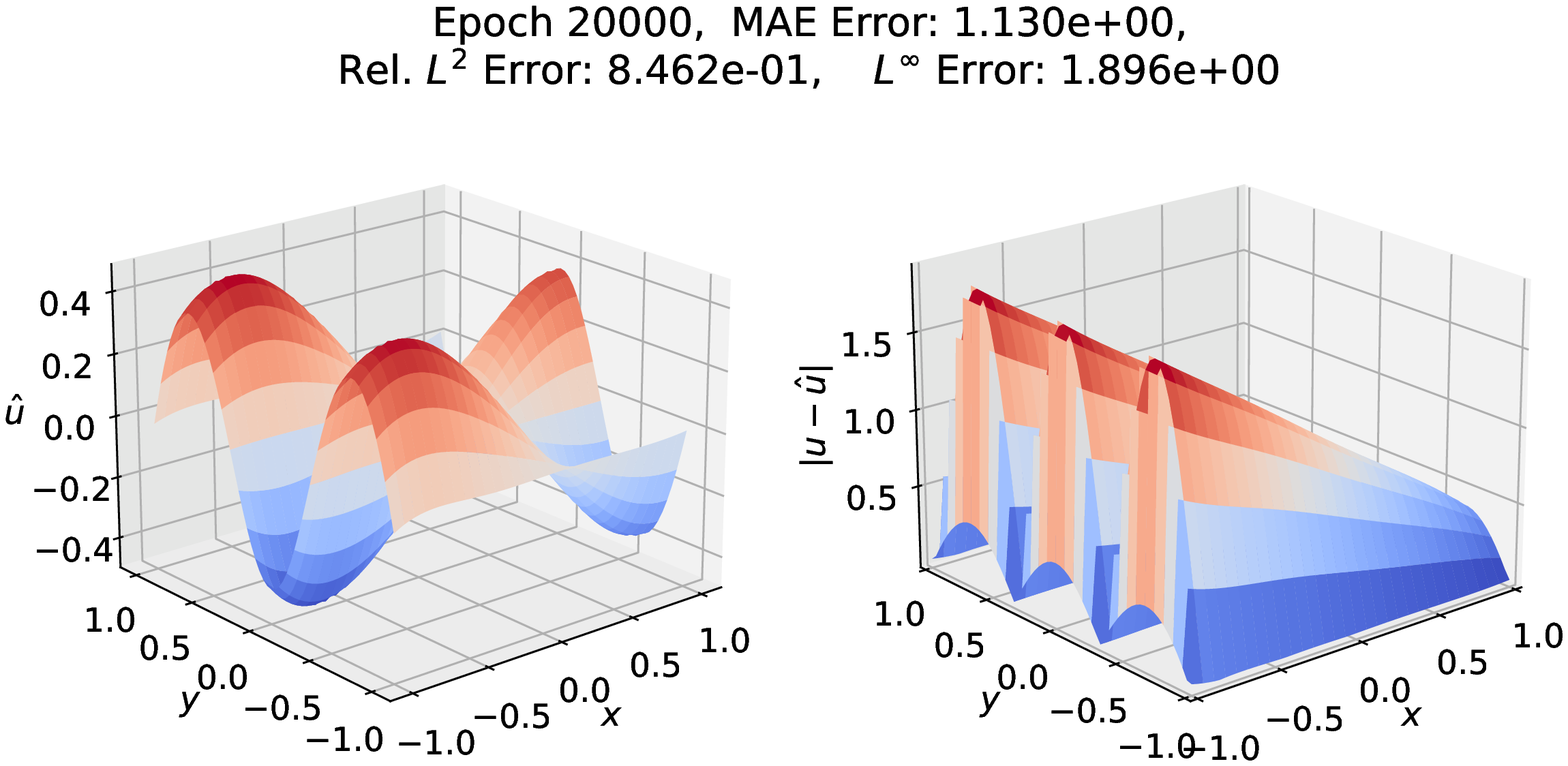}\\
     (a) Numerical experiment using the PINN method for \eqref{2D_CDEB} with $\epsilon = 10^{-6}$.
     Left: predicted solution $\widehat{u}$. Right: point-wise error between $u$ and $\widehat{u}$ where $u$ is the reference solution. \\
     \includegraphics[width=15cm]{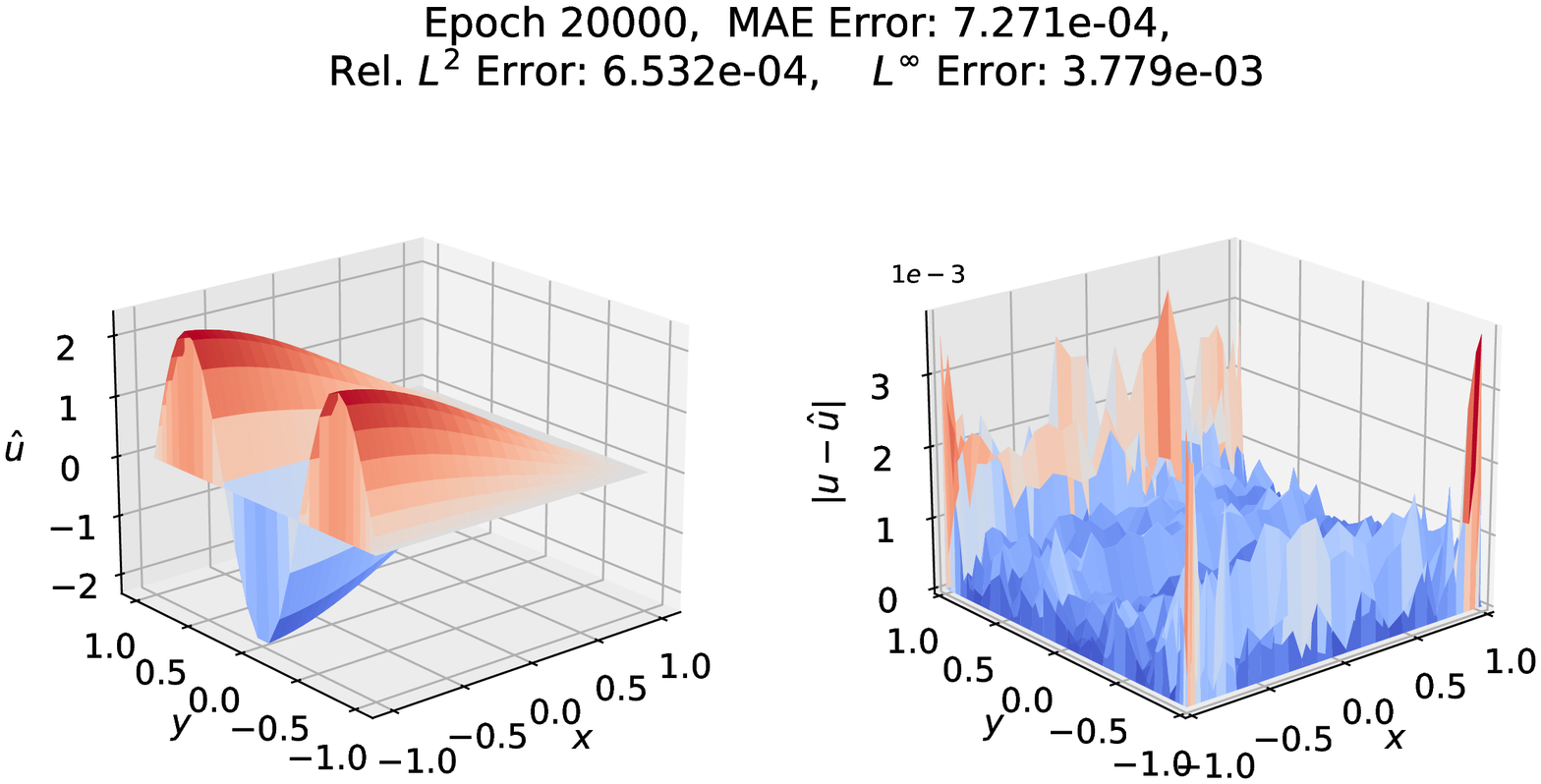}\\
     (b) Numerical experiment with ULGNet (ours) for \eqref{2D_CDEB} with $\epsilon = 10^{-6}$.
     Left: predicted solution $\widehat{u}$. Right: point-wise error between $u$ and $\widehat{u}$ where $u$ is the reference solution. 
\caption{Comparison between PINN and ULGNet (Ours) for the 2D-CDE \eqref{2D_CDEB} with $\epsilon=10^{-6}$ is displayed. 
While the PINN solution in panel (a) produces a large numerical error, the ULGNet solution in panel (b) shows an accurate numerical solution.} \label{2D_CDEB_fig}
\end{figure}
\subsection{Convection-diffusion equation with a boundary layer: two-dimensional case}
\begin{figure}[t]
\begin{tabular}{cc}
     \includegraphics[width=7cm]{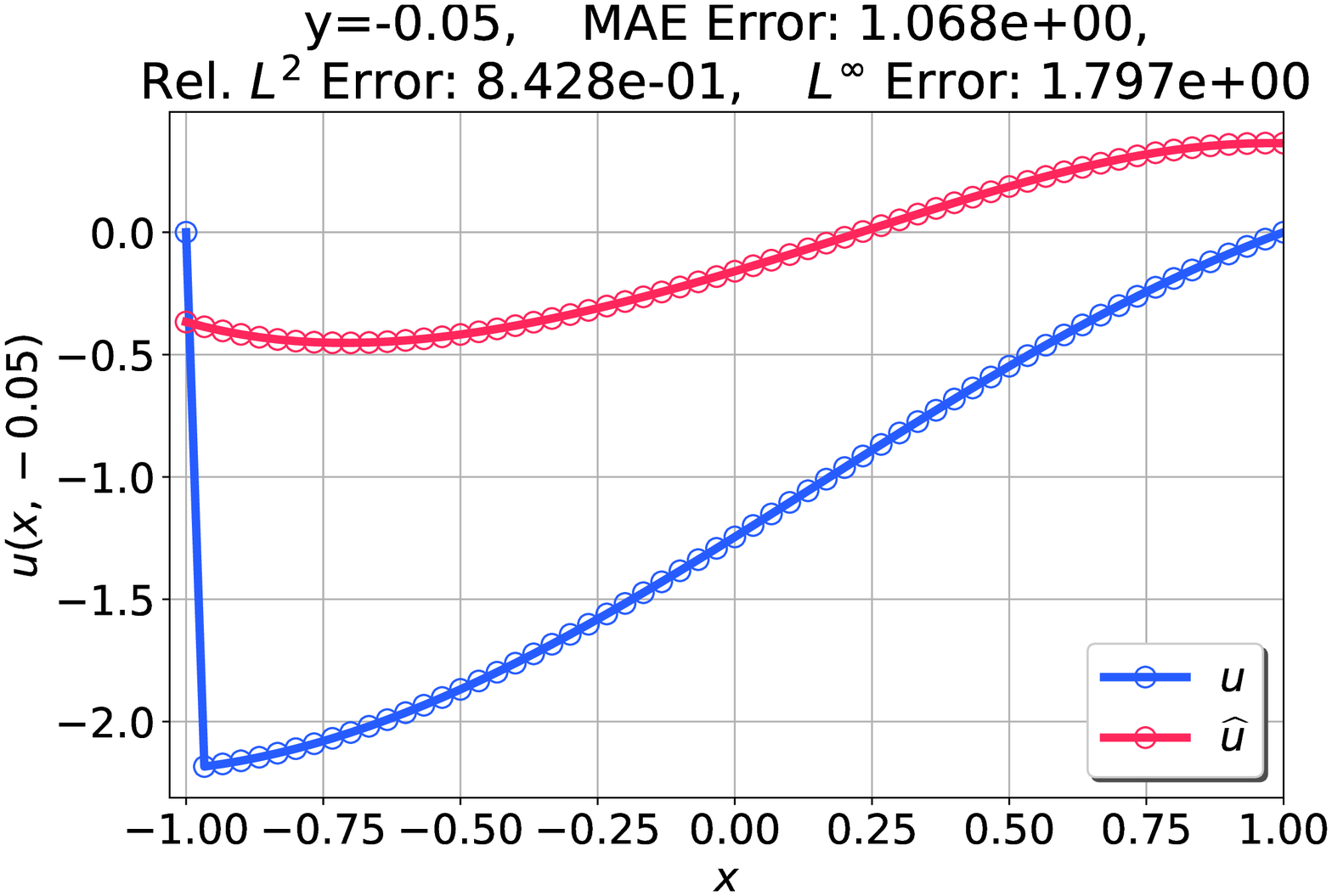}&\includegraphics[width=7cm]{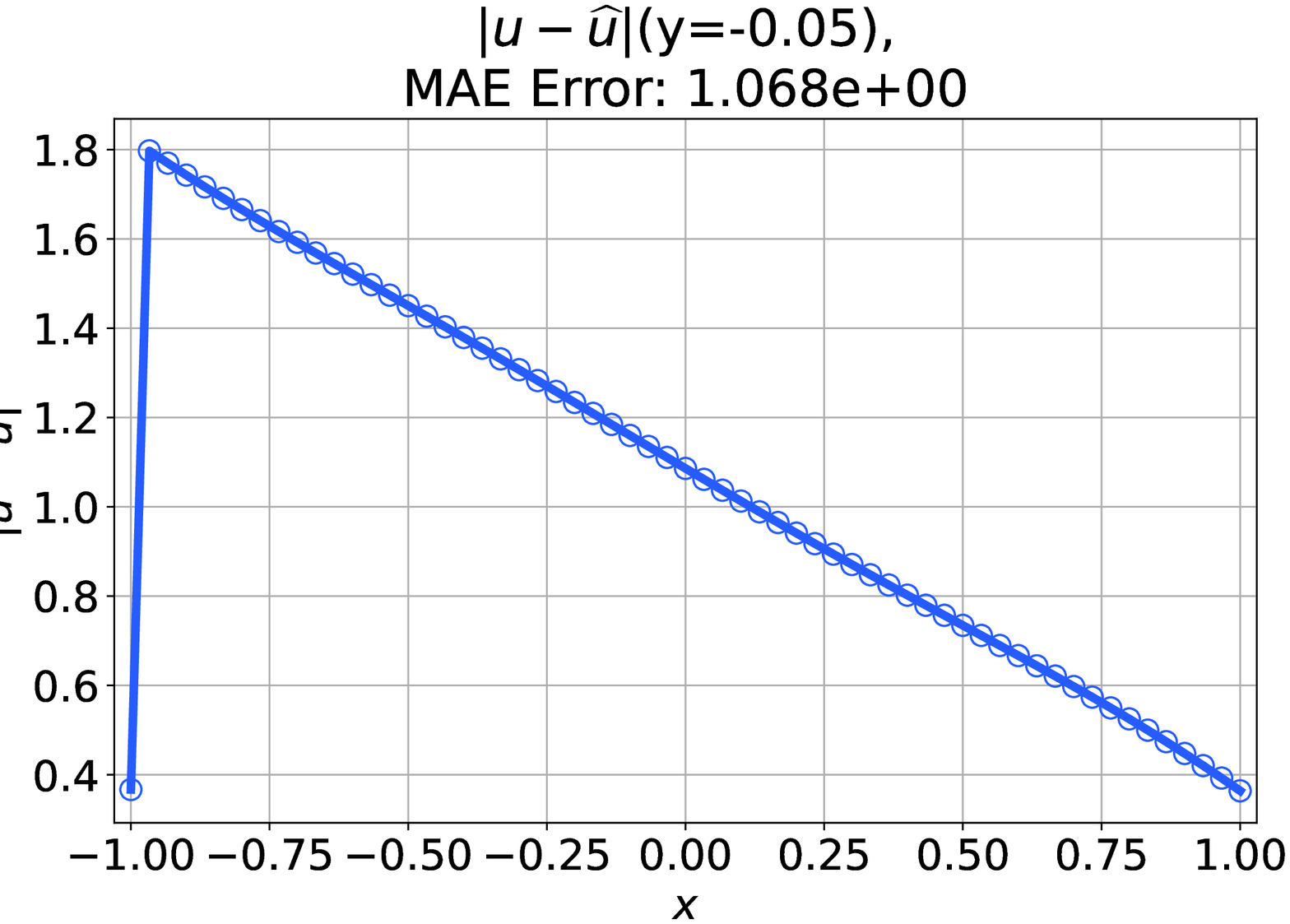}\\
     \end{tabular} \\
     {(a) The cross section of $\widehat{u}$ (PINN solution) and $u$ (reference solution) in Figure \ref{2D_CDEB_fig} (a) along the line $y = -0.05$. 
     %Numerical experiment using the PINN method.
     }\\
     \begin{tabular}{cc}
      \includegraphics[width=7cm]{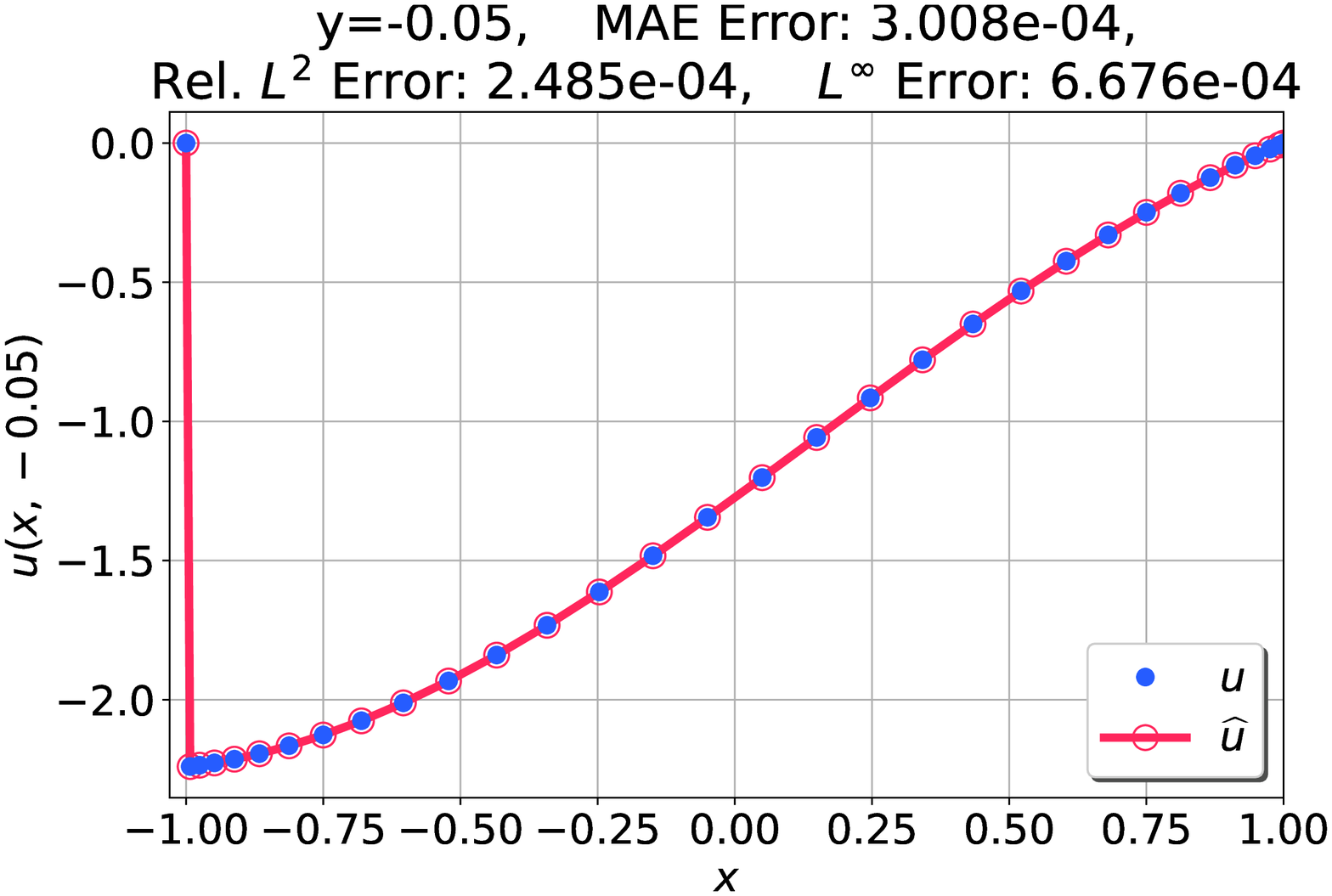}&\includegraphics[width=7cm]{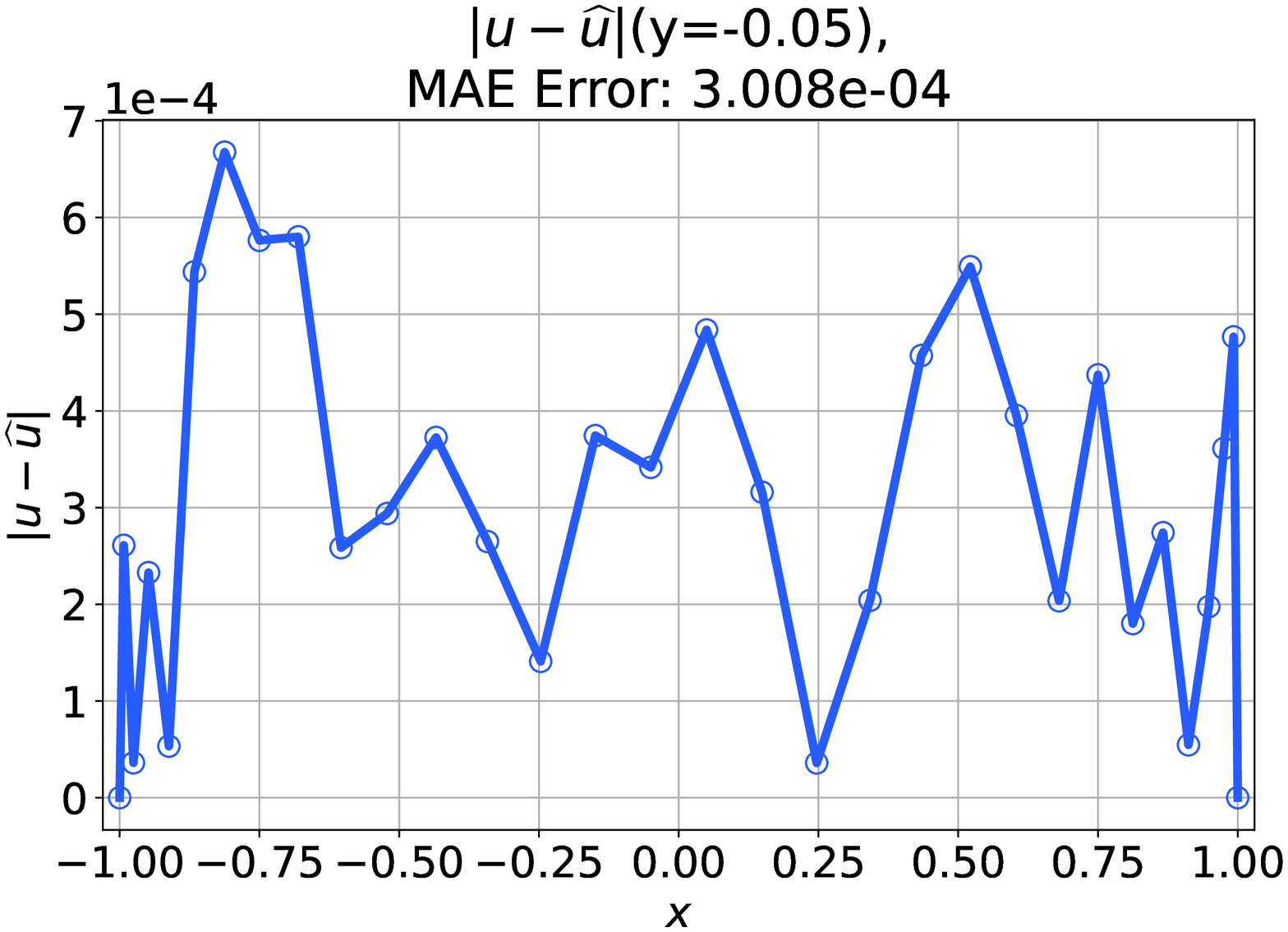}
        \end{tabular} \\
         {(b) The cross section of $\widehat{u}$ (ULGNet solution) and $u$ (reference solution) in Figure \ref{2D_CDEB_fig} (b) along the line $y = -0.05$.}
\caption{The cross section of $u$ and $\widehat{u}$ along the line $y=-0.05$ of the experiment demonstrated in Figure \ref{2D_CDEB_fig}.}\label{2D_CDEB_fig2}
\end{figure}
We further investigate the singular perturbation problem in a multi-dimension. 
Here, we consider a two dimensional convection-diffusion equation,
\begin{align}\label{2D_CDEB}
\begin{split}
&-\epsilon \Delta u+{\bf{v}}\cdot\nabla u=f(x),\quad x\in\Omega\subset\mathbb{R}^2\\
&u=0,\quad x\in\partial\Omega,
\end{split}
\end{align} 
where $\epsilon \ll 1$. 
As in \eqref{2D_CDE}, we employ $\phi_{ij}(x,y)=\phi^x_i(x)\phi^y_j(y)$ where $\phi^x_i(x)$ and $)\phi^y_j(y)$ stand for the enriched LG basis of \eqref{e:e_sum_form} in $x$ and $y$ direction, respectively.
Then, the corresponding weak equation for \eqref{2D_CDEB} is written as
\begin{align}\label{2D_CDEB_weak}
\int_{\Omega} \epsilon\nabla u\cdot\nabla\phi dx+\int_{\Omega} {\bf{v}}\cdot\nabla u \phi dx-\int_{\Omega} f\phi dx=0.
\end{align}
Accordingly, the loss function is written as 
\begin{align}\label{2D_CDE_loss}
\sum_{p=1}^{P}\sum_{i=0}^{N-1}\sum_{j=0}^{M-1}\left|\int_{\Omega} \epsilon\nabla\widehat{u}_p\cdot\nabla\phi_{ij}+ {\bf{v}}\cdot\nabla \widehat{u}_p \phi_{ij}- f_p\phi_{ij}dx\right|^2.
\end{align}
To simplify the boundary layer behavior, we assume a compatibility condition, $f=f_{yy}=0$ at $y=-1$ and $y=1$, which is required to expect the boundary layer near $x=-1$ only; see e.g., \cite{jung2005numerical}. 
We generate the inputs for \eqref{2D_CDEB} as   
\begin{align}\label{EF_2D_CDEB}
f_p(x,y)=h_{1p}\sin(\pi m_{1p}x)\sin(\frac{\pi}{2} n_{1p}(y+1))+h_{2p}\cos(\pi m_{2p}x)\sin(\frac{\pi}{2} n_{2p}(y+1)),
\end{align}
where $p=1,2,\cdots, P$, and $h_{1p}$ and $h_{2p}$ are drawn from a uniform distribution over $[1,2]$, and $m_{1p}$ and $m_{2p}$ are over $[0,2 \pi]$. In addition, $n_{1p}$ and $n_{2p}$ are integers randomly drawn from $\{1,2,3\}$. 
In order to enhance the performance, the experiments in this section are conducted with double precision floating format. 

For numerical experiments, we set $\epsilon =10^{-6}$ and ${\bf{v}}=(-1,0)$.
Figure \ref{2D_CDEB_fig} indicated that the predicted solution using the ULGNet on the out-of-sample data is close to the corresponding exact solution within MAE: $7.271e-04$, relative $L^2$: $6.532e-04$, and $L^\infty$: $3.779e-03$. 
For comparison, we provide an experiment using the PINN method in panel (a) of figure \ref{2D_CDEB_fig}.
Clearly, the PINN solution does not capture the sharp transition arose from the singular perturbation.
A closer look is available in Figure \ref{2D_CDEB_fig2}
which shows a 1D cross section of the predicted solution $\widehat{u}$ and reference solution $u$ along the line $y=-0.05$.

% \begin{figure}[t]
% \begin{tabular}{cc}
%      \includegraphics[width=7cm]{figure/pinn_loss2d.eps}&\includegraphics[width=7cm]{figure/pinn_loss_u2d.eps}\\
%      (a) Losses & (b) Relative $L^2$ error\\
%      & between the training and test sets 
%         \end{tabular}     
%      \includegraphics[width=15cm]{figure/pinn_u_upre2d.eps}\\
%      (c) Left: predicted solution $\widehat{u}$. Right: point-wise error between $u$ and $\widehat{u}$. 
% \caption{ {\color{blue} A reference experiment on the identical conditions to figure \ref{2D_CDEB_fig} but with PINN. }    }\label{Pinn2d_fig}
% \end{figure}

% \begin{figure}[t]
% \begin{tabular}{cc}
%      \includegraphics[width=7cm]{figure/cdb_cross_y1.eps}&\includegraphics[width=7cm]{figure/cdb_err_y1.eps}\\
%       \includegraphics[width=7cm]{figure/cdb_cross_y0.eps}&\includegraphics[width=7cm]{figure/cdb_err_y0.eps}\\
%       \includegraphics[width=7cm]{figure/cdb_cross_y69.eps}&\includegraphics[width=7cm]{figure/cdb_err_y69.eps}\\
%         \end{tabular} 
% \caption{Cross-sections of $u$ and $\widehat{u}$ on $y=-0.993$, $-0.050$, and $0.680$. }\label{2D_CDEB_fig}
% \end{figure}

\section{Concluding Remark}
\label{discussion}

This work proposed ULGNet which is designed to predict a solution to an elliptic PDE without access to the true solution. 
To do this, we considered the Legendre-Galerkin framework, in which the solution of a PDE can be approximated by a linear combination of a set of orthogonal bases, such as the Legendre bases, in an appropriate $L^2$ space. 
%Since the predicted solution is approximated by the spectral element scheme, the approximation process of the scheme is similar to what the proposed ULGNet is trained on. 
Hence, the approximation process of the spectral element method is similar to what the proposed ULGNet is trained on.
%Through the training, the ULGNet acquires approximations to true solutions as accurate as $10^{-4} \sim 10^{-3}$ in MAE. 
One of the main contributions of our study is the development of a learning architecture that directly tackles singular perturbation and boundary layer problems.
Thanks to the boundary layer corrector function as the additional basis function, the sharp transition from the stiff PDEs are captured by the enriched scheme.

The present research can be extended to several directions as follows. 
First, as in Section 4, the machine learning with the enriched space method can be applied to various complex problems such as interior layers Burgers equations \cite{hong2018enriched}, boundary layers on a circular domain \cite{hong2014numerical}, and 2D Navier-Stokes equations with no slip boundary condition \cite{gie2013vorticity}.
 Furthermore, since our scheme is based on the Legendre-Galerkin method, the algorithm can be extended to utilizing other orthogonal polynomials such as  Chebyshev polynomials and Jacobi polynomials depending on the features of PDEs.
 The ULGNet is also expected to be modified for finite element method (FEM) which employs, in the same manner, weak formulations of PDEs with basis functions defined on reference elements. Hence, we expect that unsupervised FEM neural network can be developed to problems on domains of various shapes.
 In future research, we also plan to apply ULGNet to time dependent problems. In terms of numerical methods, time dependent problems are implemented by Euler methods, Runge-Kutta methods, and so on. Subsequently, the methods iteratively yield the solution at the next time step from the solution at the previous time step. Thus, we expect that once ULGNet is trained on the modified time independent problem, it can iteratively generate the next time solution from the previous time solution. 

\section*{Acknowledgments}
The work of Y. H. was supported by Basic Science Research Program through the National Research Foundation of Korea (NRF) funded by the Ministry of Education (NRF-2021R1A2C1093579).

\section*{Appendix}
{\it Derivation of corrector \eqref{cor}.}\\
\\
The derivation refers to \cite{hong2018enriched}. 
We consider the CDE with small $\epsilon$,
\begin{align}\label{ACDE}
\begin{split}
&-\epsilon u^{\epsilon}_{xx}-u^{\epsilon}_x=f,\quad x\in (-1,1)\\
&u^{\epsilon}(-1)=u^{\epsilon}(1)=0.
\end{split}
\end{align}
Assuming $\epsilon=0$, the formal limit equation of \eqref{ACDE} reads
\begin{align}\label{DE0}
\begin{split}
&-u^{0}_x=f,\quad x\in (-1,1)\\
&u^{0}(1) = 0.
\end{split}
\end{align}
Because the boundary condition $u^0$ at $x=-1$ is not assigned, one can expect a discrepancy between $u^\epsilon$ and $u^0$ in the vicinity of $x=-1$. 
We observe that the boundary layer of size $\epsilon$ occurs near the outflow boundary at $x = -1$.
In this regard, we derive that the asymptotic equation for the corrector $\theta$, which approximate the discrepancy $u^\epsilon - u^0$, is given in the form,
\begin{equation} \label{ACDE}
        \begin{cases}
            -\epsilon \theta_{xx} - \theta_x = 0, 
            &
            -1 < x < 1, \\
            \theta= - u^0 , 
            &
            x = 0.
        \end{cases}
\end{equation}
The corrector $\theta$ is given in the form
\begin{equation}
    \theta(x)
        =
            -u^0(0) \, e^{-x/\epsilon}
            + e.s.t., 
\end{equation}
where the $e.s.t.$ denotes an exponentially small term with respect to $\epsilon$. 
The 2D boundary corrector can be derived in a similar fashion to the 1D corrector; for more details, see e.g., \cite{hong2018enriched,jung2005numerical}.

\bibliographystyle{unsrtnat} 
\bibliography{ULGNet_ref}  %%% Uncomment this line and comment out the ``thebibliography'' section below to use the external .bib file (using bibtex) .





% \end{thebibliography}

\end{document}